\documentclass[journal]{IEEEtran}
\usepackage{lineno,hyperref}
\usepackage{bbm}
\usepackage{bm}
\usepackage{amsfonts}
\usepackage{amsmath}
\usepackage{mathrsfs}
\usepackage{amssymb}
\usepackage{enumerate}
\usepackage{graphicx}
\usepackage{subfigure}
\usepackage{booktabs}
\usepackage{graphicx}
\usepackage{setspace}
\usepackage{algorithm}
\usepackage{algorithmic}
\usepackage{setspace}
\usepackage{multirow}
\usepackage{color}
\usepackage{cite}
\usepackage{array}
\usepackage{times}
\usepackage{mathptmx}
\usepackage{dsfont}
\usepackage{float}
\ifCLASSINFOpdf
\else
\fi
\begin{document}
\title{Generation and Recombination for Multifocus Image Fusion with Free Number of Inputs}
\author{Huafeng~Li, Dan~Wang, Yuxin Huang, Yafei Zhang and Zhengtao~Yu\IEEEmembership{}
\thanks{This work was supported in part by the National Natural Science Foundation of China (61966021, 62276120, 62161015), and the Yunnan Fundamental Research Projects (202301AV070004).}
\thanks{H. Li, D. Wang, X. Huang, Yafei Zhang and Z. Yu are with the Faculty of Information Engineering and Automation, Kunming University of Science and Technology, Kunming 650500, China.(E-mail:lhfchina99@kust.edu.cn (H. Li); wd97@kust.edu.cn (D. Wang))}
\thanks{}
\thanks{Manuscript received xxxx;}}
\markboth{Journal of \LaTeX\ Class Files}%
{Shell \MakeLowercase{\textit{et al.}}}
\maketitle
\begin{abstract}
Multifocus image fusion is an effective way to overcome the limitation of optical lenses. Many existing methods obtain fused results by generating decision maps. However, such methods often assume that the focused areas of the two source images are complementary, making it impossible to achieve simultaneous fusion of multiple images. Additionally, the existing methods ignore the impact of hard pixels on fusion performance, limiting the visual quality improvement of fusion image. To address these issues, a combining generation and recombination model, termed as GRFusion, is proposed. In GRFusion, focus property detection of each source image can be implemented independently, enabling simultaneous fusion of multiple source images and avoiding information loss caused by alternating fusion. This makes GRFusion free from the number of inputs. To distinguish the hard pixels from the source images, we achieve the determination of hard pixels by considering the inconsistency among the detection results of focus areas in source images. Furthermore, a multi-directional gradient embedding method for generating full focus images is proposed. Subsequently, a hard-pixel-guided recombination mechanism for constructing fused result is devised, effectively integrating the complementary advantages of feature reconstruction-based method and focused pixel recombination-based method. Extensive experimental results demonstrate the effectiveness and the superiority of the proposed method. The source code will be released on \url{https://github.com/xxx/xxx}.

\end{abstract}
\begin{IEEEkeywords}
Multifocus image fusion; Image generation; Free number of inputs; Hard pixel detection.
\end{IEEEkeywords}
\IEEEpeerreviewmaketitle

\section{Introduction}
Multifocus image fusion aims to integrate the information of several images captured from the same scene with different focus settings into a single image where all regions are in focus. This technique has received  wide attention from researchers as it could overcome the depth of field limitation of optical lenses and produce an image with all objects in focus. Roughly, the existing fusion methods can be classified into two main categories: feature reconstruction-based  methods and focused pixel recombination-based methods.

Feature reconstruction-based methods extract features from the source images as the initial step, followed by fusing these features using specific fusion strategies. Finally, the fused features are used to generate the fusion result. Typically, feature extraction methods include multiscale geometric analysis methods(such as wavelet transform \cite{2}, multiscale contourlet transform \cite{50}) and deep neural networks \cite{52, 53}, \textit{etc}. Fused images obtained by feature reconstruction-based methods often exhibit high visual quality and can effectively avoid block artifacts. However, since the fusion result is reconstructed or generated based on fused features, there is still a slight gap between the fusion image and the ideal result. Focused pixel recombination-based methods typically involve detection and recombination of focused pixels. Although these methods usually achieve better objective evaluation quality, they are prone to introducing artifacts due to ambiguity of pixel focus property, resulting in the degradation of the visual quality of the fusion result.
\begin{figure}[t!]
	\centering
	\includegraphics[width=3.0in,height=2.45in]{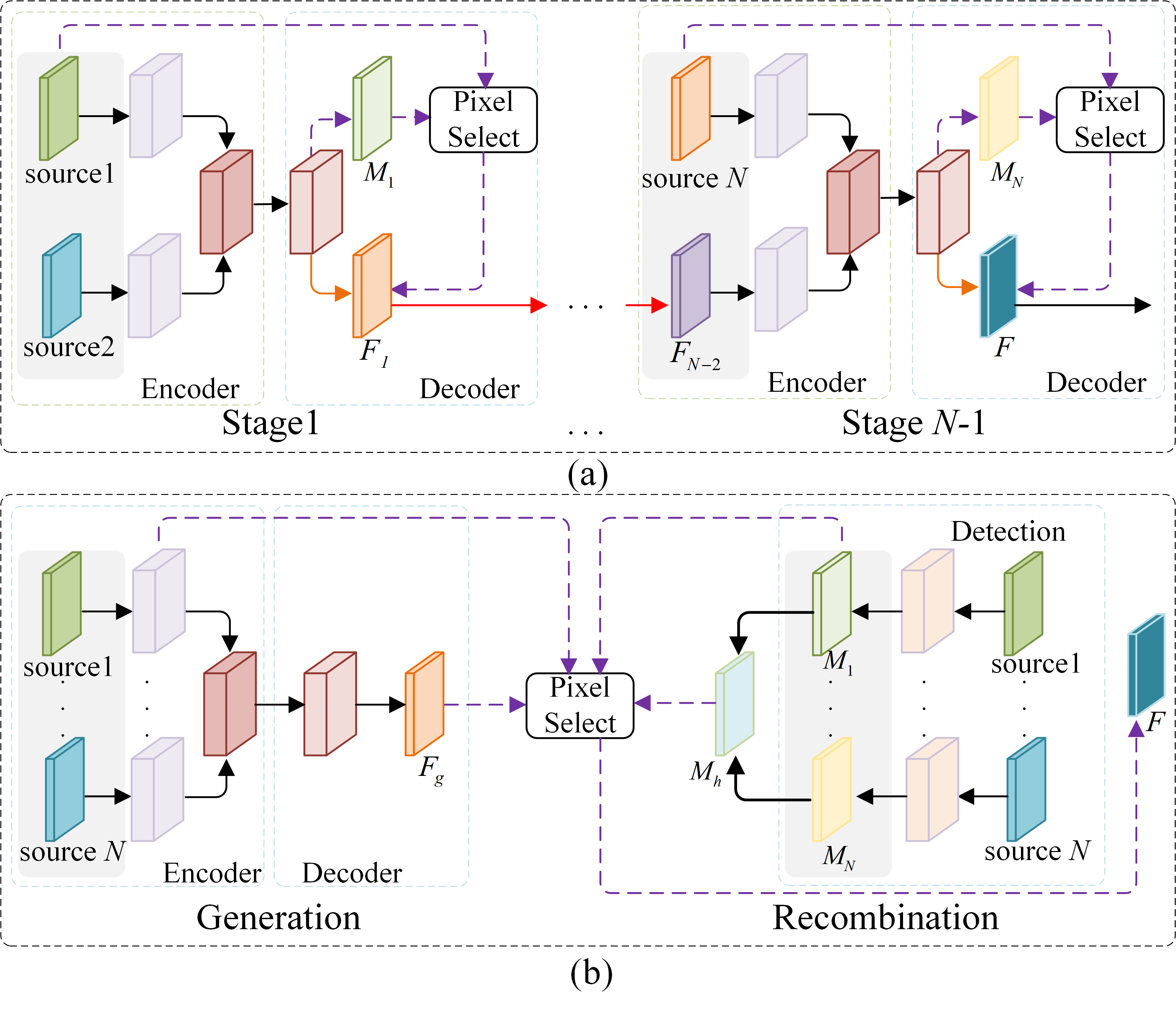}
	\setlength{\abovecaptionskip}{0.cm}
	\caption{Multi-image fusion strategy for multiple images. (a) Alternating fusion strategy used in existing methods. (b) Fusion strategy in this paper.}
	\label{label}
	\vspace{-0.6cm}
\end{figure}

To integrate the advantages of these two types of methods, Liu \textit{et al}.\cite{25} proposed to combine deep residual learning and focus property detection to achieve the fusion of multifocus images. This method assumes that the focused regions between the two source images are complementary. Based on this assumption, the detection results of focused pixels in one source image can be obtained immediately after the focus property of another source image is determined. It's a commonly used method to obtain focus property detection of two source images. However, this strategy has the following limitations. Firstly, there are a large number of hard pixels in the source images, which exhibits significant ambiguity in their focus properties. These pixels typically come from the smooth regions or boundaries of focused regions. It's difficult to determine whether the pixels are in focus or not, so there will be a large number of misclassifications in detection result of focused pixels. Subsequently, the quality of the fused image will be compromised if the above detection result is directly used to determine the focus property of another source image.
Secondly, these methods can only fuse two images at a time. When faced with the fusion of multiple images, they typically adopt an alternating fusion strategy as shown in Fig.\ref{label} (a), \textit{i.e}., fuse two source images firstly, and then fuse the fused result of the previous stage with that of the remaining source images, until all source images are fused. Although the strategy is feasible, it may lead to information loss due to multiple feature extraction and fusion, which is detrimental to the final fused image.

To address the aforementioned issues, we propose a multifocus image fusion method named GRFusion that is open to the number of fused images. Specifically, the focus property of each source image is determined independently with the assistance of remaining source images. Therefore, the detection result of one source image will no longer affect that of other images. Furthermore, owing to the independent focus property detection for each source image, GRFusion is endowed with the ability to process the fusion of arbitrary number of images in an open manner, reducing the risk of information loss caused by alternating fusion. Additionally, to integrate the complementary advantages of feature reconstruction-based and pixel recombination-based methods, a hard-pixel-guided fusion result construction mechanism is devised. It can effectively reduce the negative impact caused by hard pixels on the fusion result. Technically, this method can determine the origin of the corresponding pixel in the fused image based on whether the focus property of pixel is determined or not. Specifically, for the pixel whose focus property is determined, the corresponding pixel in fused result is derived from the corresponding source image. Otherwise, the pixel in fused image is derived from the result generated by a feature reconstruction-based method with multi-directional gradient embedding proposed in this paper. In summary, this paper has three main contributions as follows.
\begin{itemize}
\item A hard pixels detection method is designed to meet the requirement of parallel fusion of multiple images, where the hard pixels are identified by finding outliers from the detection results of the focus property of the source image. Owing to the independent pixel focus property detection, the developed method is free from the number of fused source images.

By integrating the advantages of feature reconstruction-based method and pixel recombination-based method into a single fusion framework, a novel multifocus image fusion approach termed GRFusion is devised. Unlike existing methods that only use this combined fusion strategy in the transition region between focused and defocused regions. GRFusion applies this strategy to all hard pixels, thus avoiding the negative impact on fused result caused by the detection error of pixel focus property.

\item Experimental results on both visual quality and objective evaluation demonstrate the effectiveness and superiority of our method in comparison to 9 well-known multifocus image fusion methods.
\end{itemize}

The rest of the paper is structured as follows: Section \ref{section1} briefly reviews the related work; Section \ref{section3} describes the details of the proposed method; Section \ref{section4} gives the analysis of the experimental results; And Section \ref{section5} summarizes the paper and draws some conclusions.

\section{Related Work}\label{section1}
\subsection{Feature Reconstruction-Based Multifocus Image Fusion}
Feature reconstruction-based methods usually fuse the extracted multifocus image features and then generate the fused images based on the fused features. In such methods, the representation of image features is a crucial factor that influences the fusion image. Currently, the commonly used methods for image feature representation include multi-scale decomposition (MSD) \cite{1,2,4,6,46}, sparse representation (SR)\cite{47,7,8,9,10,11,12,49}, and deep learning (DL)\cite{13, 53, 14,15,16,17,54, 55}. MSD-based methods often employ multiscale image analysis method  such as Wavelet Transform\cite{1,2}, Nonsubsampled Contourlet Transform\cite{4, 50}, and Curvelet Transform\cite{6} to represent the the source images. However, these methods employ fixed basis to represent the input image, so their sparse representation ability is weak. SR-based methods have been widely used in multifocus image fusion since they can learn a set of basis from the training samples to sparsely represent the image. Specifically, Yang \textit{et al}.\cite{10} proposed a SR-based method for restoration and fusion of noisy multifocus images. Zhang \textit{et al}.\cite{11} developed an analysis-synthesis dictionary pair learning method for fusion of multi-source images, including multifocus images. Li \textit{et al}.\cite{12} devised an edge-preserving method for fusion of multi-source noisy images. This method can perform fusion and denoising of multifocus noisy images in one framework.
\begin{figure*}[t!]
	\centering
	\includegraphics[width=6.6in,height=3.6in]{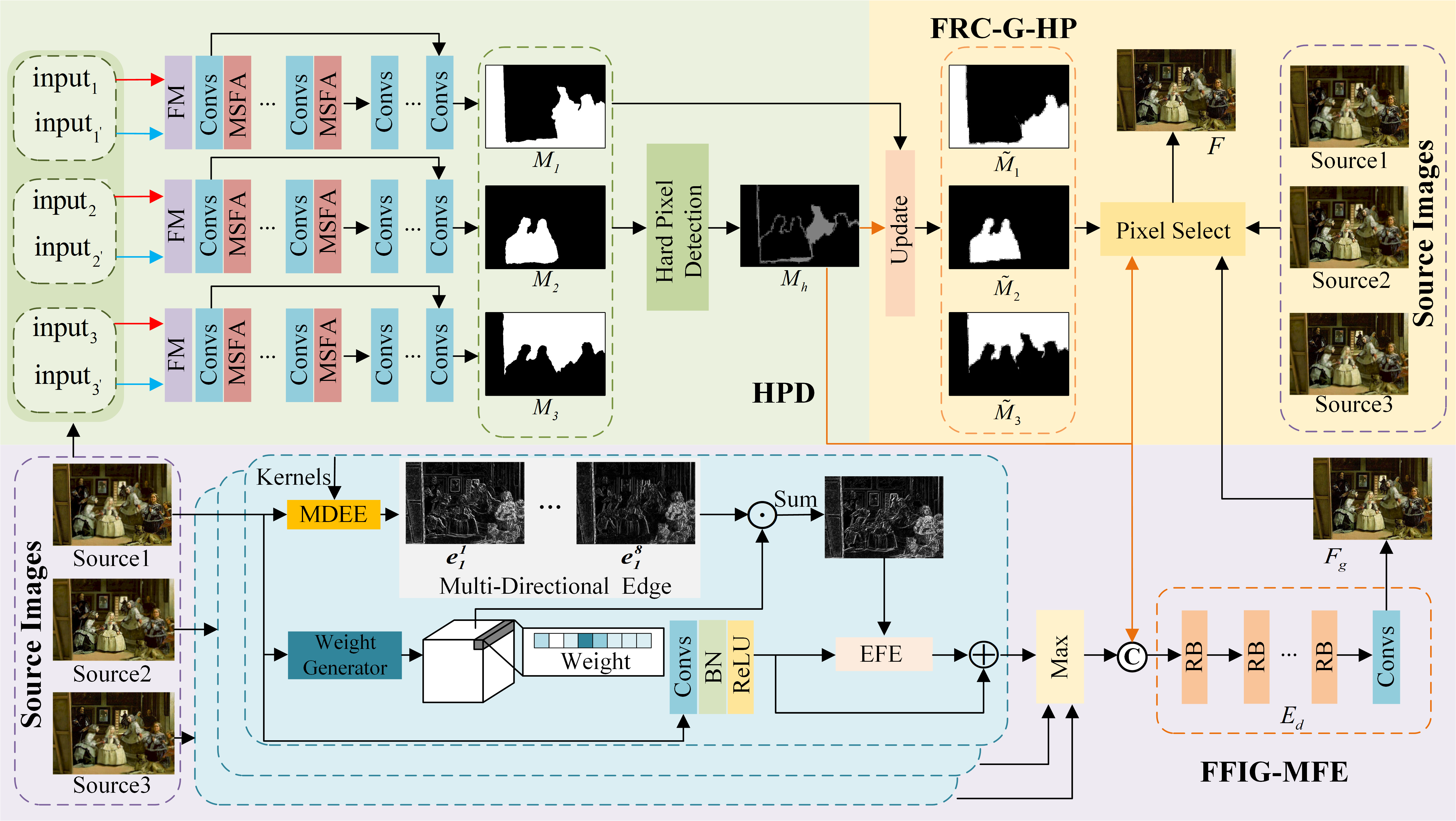}
	\setlength{\abovecaptionskip}{0.cm}
	\caption{Overall framework of the proposed method. Here we take the fusion of three images as an example. In the proposed method, under the guidance of the remaining images, each image can independently detect focus property of pixel, so the method is free from the number of input images.}
	\label{labe2}
	\vspace{-0.6cm}
\end{figure*}

In DL-based methods, the feature reconstruction-based approach generally includes three steps: feature extraction, feature fusion, and fusion image reconstruction. Since the fusion result is reconstructed from fused features, there may be some errors between the reconstructed and the ideal results. This demonstrates that the result obtained by the method based on feature reconstruction is not ideal.  To alleviate this issue, Zhao \textit{et al}.\cite{13} proposed a CNN-based multi-level supervised network for multifocus image fusion, which utilized multi-level supervision to refine the extracted features. Zang \textit{et al}.\cite{14} introduced spatial and channel attention to enhance the salient information of the source images, effectively avoiding the loss of details. To address the challenge raised by the absence of ground truth, Jung \textit{et al}.\cite{15} proposed an unsupervised CNN framework for fusion of multi-source images, where an unsupervised loss function was deigned for unsupervised training.

In an unsupervised learning mode, to prevent the loss of source image details, Mustafa \textit{et al}.\cite{58} added dense connections into feature extraction network at different levels, improving the visual quality of the fused images. Hu \textit{et al}.\cite{17} proposed a mask prior network and a deep image prior network to achieve multifocus image fusion with zero-shot learning, thereby solving the problem caused by inconsistency between synthetic images and real images. Although the feature reconstruction-based method can generate images that look natural, there is still a certain gap between the generated and the ideal results due to the information loss during the feature extraction and image reconstruction.

\subsection{Pixel Recombination-Based Multifocus Image Fusion}
Pixel recombination-based methods usually combine the focused pixels of the source image to construct the fused image. Therefore, how to accurately detect the pixels in the focus area is critical to them. To this end, Liu \textit{et al}.\cite{18} proposed a twin network and trained with a large number of samples generated by gaussian filtering to achieve the detection of focused pixels. Guo \textit{et al}.\cite{19} employed a fully connected network to explore the relationship between pixels in the entire image, enabling the determination of pixel focus property. Ma \textit{et al}.\cite{20} introduced an encoder-decoder feature extraction network consisting of SEDense Blocks to extract the features from source images, and then the focused pixel is distinguished by measuring spatial frequency of the features. Xiao \textit{et al}.\cite{21} proposed a U-Net with global-feature encoding to determine the focus property of pixel. Ma \textit{et al}.\cite{22} proposed SMFuse for focused pixel detection where the repeated blur map and the guided filter are both used to generate the binary mask. Considering the complementarity of features at different receptive fields, Liu \textit{et al}.\cite{23} proposed multiscale feature interactive network to improve detection performance of the focused pixel.

The aforementioned methods usually construct fused image using the detected focused pixels. However, it is extremely challenging to determine the focus property of hard pixels located at the smooth regions or the boundaries of focused regions. If the fused image is obtained like above, severe artifacts will be introduced into the fused result. To solve this problem, Li \textit{et al}.\cite{24} proposed an effective multifocus image fusion method based on non-local means filtering, in which the detection result of focused pixel in the transition regions between focused and defocused areas is smoothed by the non-local means filtering, and then the smoothed result is used to guide the fusion process. Similarly, Liu \textit{et al}.\cite{25} combined feature reconstruction with focused pixel recombination, and then developed an effective method for fusion of multifocus images. Although above methods are effective, they ignore the negative impact of hard pixels on the fusion quality. Additionally, the most of existing methods often assume that the number of input images is two and the focus property is complementary. As a result, they cannot fuse the multiple images in a parallel mode. Alternating fusion strategy is a commonly used method to solve this problem, but it tends to lose the information of the source images. In contrast, the developed GRFusion can fuse multiple images simultaneously and can solve the problems caused by hard pixels effectively.
\section{Proposed Method}\label{section3}
\subsection{Overview}
As shown in Fig. \ref{labe2}, the proposed method mainly consists of three components: hard pixel detection (HPD), full-focus image generation with multi-directional feature embedding (FFIG-MFE), and fusion result construction guided by hard pixels (FRC-G-HP). HPD is mainly used to determine which pixels are from the focus ambiguity area. This allows us to take specific solution to address the fusion of hard pixels. FFIG-MFE is used to generate an all-focus fusion image based on the fused features of the source images. FRC-G-HP combines the all-focus fusion result and the source images to generate the final fused result under the guidance of detection results of focused pixels. In the following section, we will elaborate on these components carefully.
\subsection{Hard Pixel Detection}
\subsubsection{Focus Modulation}
\begin{figure}[t!]
	\centering
	\includegraphics[width=3.4in,height=2.0in]{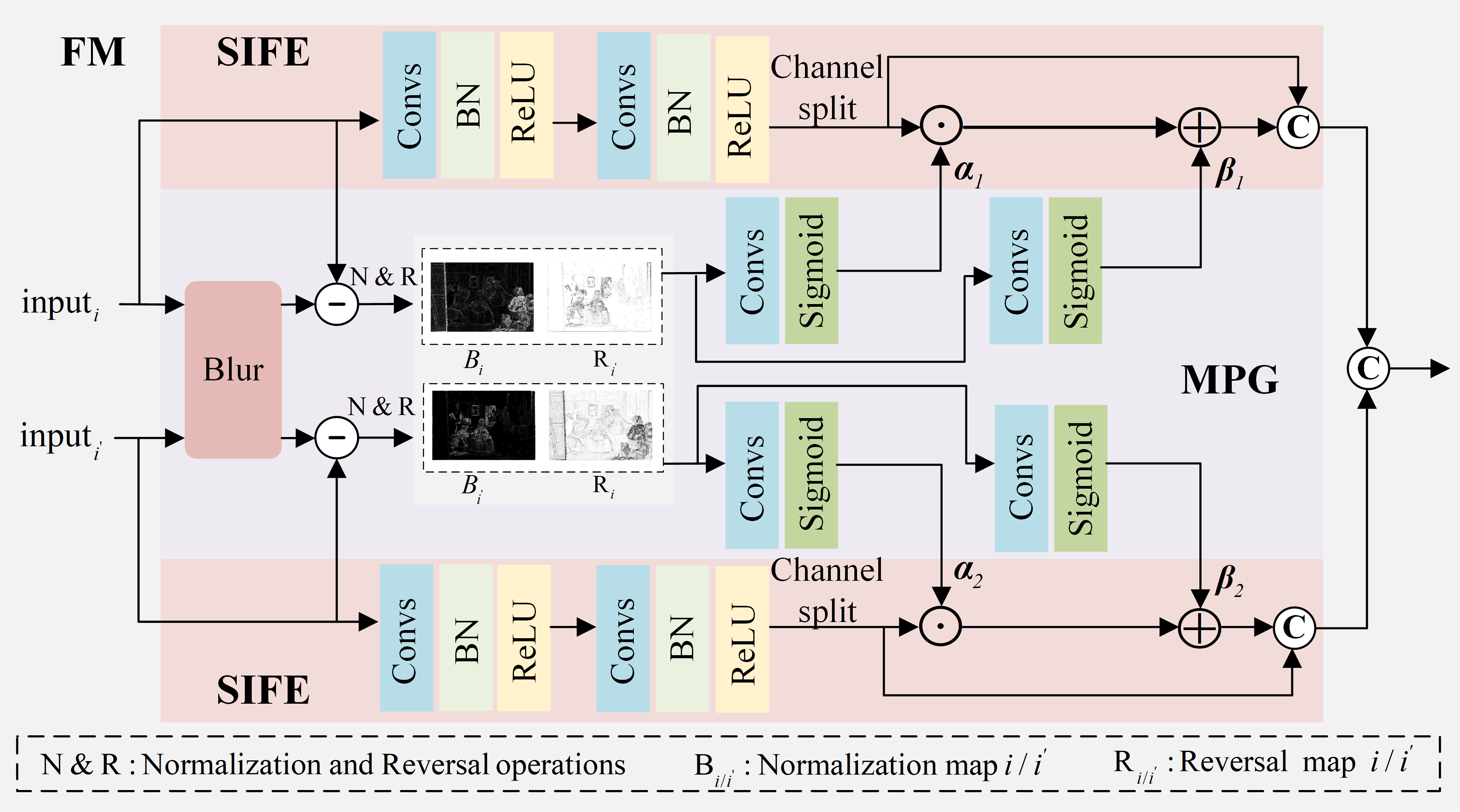}
	\setlength{\abovecaptionskip}{0.cm}
	\caption{The detailed architecture of FM.}
	\label{labe3}
	\vspace{-0.6cm}
\end{figure}

As shown in Fig.\ref{labe2}, the HPD is mainly composed of two sub-modules: focus modulation (FM) and multi-scale feature aggregation (MSFA). Meanwhile, convolutional layers are embedded between FM and MSFA to further extract features. FM is mainly used to adjust the discriminability of focused features and improve the detection accuracy of pixel focus property. The specific process is shown in Fig.\ref{labe3}. It consists of source image feature extractor (SIFE) and modulation parameter generator (MPG). Let $inpu{t_i}$ and $inpu{t_{i'}}$ be the inputs of FM,  where $inpu{t_i}$ denotes the $i$-th source image, and $inpu{t_{i'}}$ denotes the sum of remaining source images. $inpu{t_{i'}}$ is mainly used to provide assistance for the focus detection of $inpu{t_i}$.  SIFE is used to extract features from $inpu{t_i}$ and $inpu{t_{i'}}$, and it consists of convolutional operation, batch normalization and ReLU activation function. MPG is used to generate the modulation parameters to highlight the effect of important features. Technically, we first obtain the edge details of two inputs by:
\begin{equation}
	\begin{aligned}
		edg{e_l} = inpu{t_l} - inpu{t_l}\circledast{\bm G} (l=i, i')
	\end{aligned},
\end{equation}
where $\bm G$ denotes the Gaussian blur kernel and $\circledast$ denotes the convolutional operation. In order to highlight the focus property of each pixel in the source images and prevent the data in $\bm edg{e_l}$ from being dispersed, a normalization process is performed on $edg{e_l}$:
\begin{equation}
\begin{aligned}
{\bm B_l} = \frac{{edg{e_l} - \min (edg{e_l})}}{{\max (edg{e_l}) - \min (edg{e_l})}} {\rm{(}}l{\rm{=}}i{\rm{,}}i'{\rm{)}}
\end{aligned}.
\end{equation}

To make full use of the focus information in $\bm edg{e_l}$, we reverse the normalization result to obtain:
\begin{equation}
	\begin{aligned}
	{\bm R_l} = {\bf{1}} - \bm B_{l}		
	\end{aligned},
\end{equation}
where $\bm {1}$ denotes an all-one matrix with the same size as ${\bm R_l}$. Since ${\bm R_i}$ and ${\bm B_{i'}}$ share the same focus property, the concatenated result can highlight the effect of the feature in focus region.
For this reason, we send the concatenated features $[{\bm B_i},{\bm R_{i'}}]$ to two sub-networks which both consist of $3 \times 3$ convolution and $sigmoid$ activation function to generate scale factor ${\bm \alpha _i} $ and transform parameter ${\bm \beta _i}$, respectively:
\begin{equation}
	\begin{aligned}
	{\bm \alpha _i} = {\rm{sigmoid}}(Con{v_{3 \times 3}}([{\bm B_i},{\bm R_{i'}}])) \\
	{\bm \beta _i} = {\rm{sigmoid}}(Con{v_{3 \times 3}}([{\bm B_i},{\bm R_{i'}}])) \\
\end{aligned},
\end{equation}
where $[\cdot,\cdot]$ denotes concatenation operation along the channel dimension, $Con{v_{3 \times 3}}$ denotes $3 \times 3$ convolutional layer. It should be noted that the parameters in the two convolutional layer in Eq.(4) are not shared.

To prevent the over-modulation of ${\bm \alpha _i}$ and ${\bm \beta _i}$ from causing irreparable effects on the features, the parameters ${\bm \alpha _i}$ and ${\bm \beta _i}$ are only performed on the first half of the feature channels of $inpu{t_l} (l = i, i')$. In detail, $inpu{t_l}(l = i,i')$ is sent into a feature extraction network consisting of $3 \times 3$ convolutional layer, BN and ReLU activation function, and then the results obtained after channel split can be expressed as:
\begin{equation}
\begin{aligned}
(\overleftarrow{\bm F_i}, \overrightarrow{\bm F_i}) = {\rm split}(ReLU{(BN(Conv(input_i,3 \times 3)))_{ \times 2}})
\end{aligned},
\end{equation}
where ${\overleftarrow{\bm F_i}}$ and ${\overrightarrow{\bm F_i}}$ denote the first half features and the second half features of $inpu{t_l}(l = i,i')$ along the channel, respectively. The result after ${\bm \alpha _i} $ and ${\bm \beta _i} $ modulation is:
\begin{equation}
	\begin{aligned}
		\bm F_i^{\text{mod}} = [{\bm \alpha _i} \odot \bm {\overleftarrow{\bm F_i}} + {\bm \beta _i},\bm {\overrightarrow{\bm F_i}}]
	\end{aligned}.
\end{equation}
$\bm F_i^{\text{mod}}$ denotes the result after focus modulation. To use $\bm F_i^{\text{mod}}$ and $\bm F_{i'}^{\text{mod}}$ jointly, in this paper, $\bm F_i^{\text{mod}}$ and $\bm F_{i'}^{\text{mod}}$ are concatenated and then fed into MSFA to further extract the features of focused pixels.

\subsubsection{Multi-Scale Feature Aggregation}
\begin{figure}[t!]
	\centering
	\includegraphics[width=3.4in,height=2.0in]{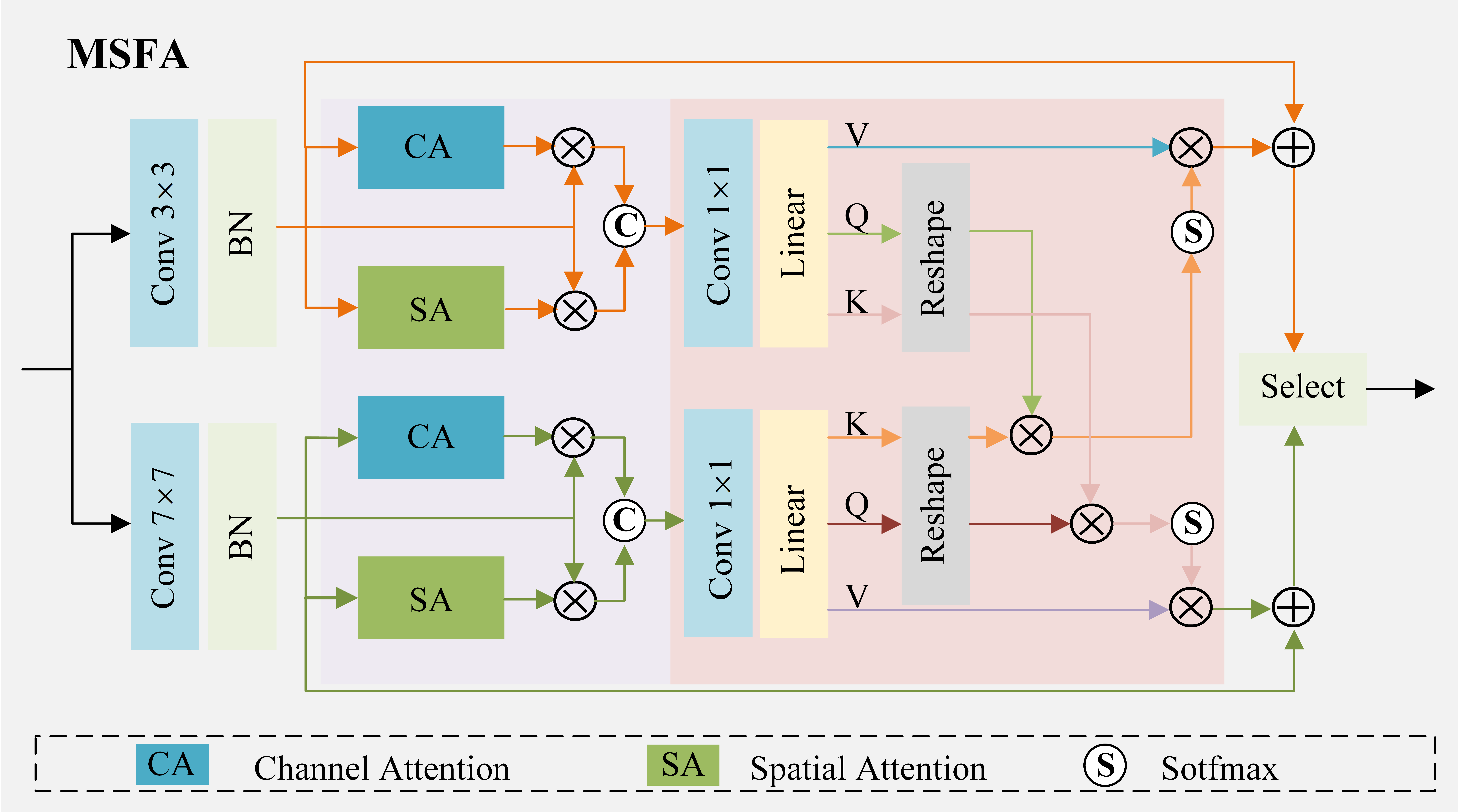}
	\setlength{\abovecaptionskip}{0.cm}
	\caption{The detailed architecture of MSFA.}
	\label{labe4}
	\vspace{-0.6cm}
\end{figure}
After FM, the features in the focused region are highlighted but still not enough to make a correct determination for the focus property of pixels. To this end, MSFA is designed as shown in Fig.\ref{labe4}. In this module, the saliency of the features in the focused region is improved by interactive enhancement of features extracted under different receptive fields. Methodologically, MSFA mainly consists of a multiscale convolution, BN, spatial attention (SA)\cite{37}, channel attention(CA) \cite{37}, $1 \times 1$ convolution, and multiscale cross-attention. In detail, we send $[\bm F_{i}^{\text{mod}}, \bm F_{i'}^{\text{mod}}]$ into MSFA, the outputs of BN can be formulated:
\begin{equation}
\begin{array}{l}
	\bm F_i^{3 \times 3} = BN(Con{v_{3 \times 3}}([\bm F_{i}^{\text{mod}}, \bm F_{i'}^{\text{mod}}]) \\
	\bm F_i^{7 \times 7} = BN(Con{v_{7 \times 7}}([\bm F_{i}^{\text{mod}}, \bm F_{i'}^{\text{mod}}]) \\
\end{array}.
\end{equation}
To highlight the effect of important information in $\bm F_i^{3 \times 3}$ and $\bm F_i^{7 \times 7}$ in hard pixel detection,  $\bm F_i^{3 \times 3}$ and $\bm F_i^{7 \times 7}$ are sent into SA and CA and then integrated by ${1 \times 1}$ convolution:
\begin{equation}
	\begin{array}{l}
		\bm{\bar{F}}_i^{3 \times 3} = {{Con}}{{\rm{v}}_{1 \times 1}}[CA(\bm{F}_i^{3 \times 3}),SA(\bm{F}_i^{3 \times 3})] \\
		\bm{\bar{F}}_i^{7 \times 7} = {{Con}}{{\rm{v}}_{1 \times 1}}[CA(\bm F_i^{7 \times 7}),SA(\bm F_i^{7 \times 7})] \\
	\end{array}.
\end{equation}

Features under different perspective fields not only have consistent focus property, but also have certain complementary effects in focus property detection.
If the features of the same image at different scales can be used to assist each other and enhance the focus information, the saliency of the focus area features will be effectively highlighted\cite{56}, which is beneficial for determining the pixel focus property. To this end, a feature saliency cross-enhancement mechanism is introduced into MSFA. Specifically, $\bm{\tilde {F}}_i^{3 \times 3}$ and $\bm{\tilde{F}}_i^{7 \times 7}$ are obtained after linearly mapping of $\bm{\bar{F}}_i^{3 \times 3}$ and $\bm{\bar{F}}_i^{7 \times 7}$:
\begin{equation}
	\begin{array}{l}
		\bm{\tilde {F}}_{Q,i}^{3 \times 3} = \bm {W}_Q^3 \bm{\bar{F}}_i^{3 \times 3},
		\bm{\tilde {F}}_{K,i}^{3 \times 3} = \bm {W}_K^3 \bm{\bar{F}}_i^{3 \times 3},
		\bm{\tilde {F}}_{V,i}^{3 \times 3} = \bm {W}_V^3 \bm{\bar{F}}_i^{3 \times 3} \\
		\bm{\tilde {F}}_{Q,i}^{7 \times 7} = \bm {W}_Q^7 \bm{\bar{F}}_i^{7 \times 7},
		\bm{\tilde {F}}_{K,i}^{7 \times 7} = \bm {W}_K^7 \bm{\bar{F}}_i^{7 \times 7},
		\bm{\tilde {F}}_{V,i}^{7 \times 7} = \bm {W}_V^7 \bm{\bar{F}}_i^{7 \times 7} \\
	\end{array},
\end{equation}
where $\bm W_{l'}^{k}(l' = \bm Q, \bm K, \bm V; k = 3,7)$ is a linear transformation matrix. The dimension of $\bm{\tilde {F}}_{l',i}^k \in {\mathbb{R}^{C \times H \times W}}(l' = Q,K;k = 3,7)$ is changed to $C \times (HW)$ after reshaping operation. To enhance the features of focused region, the feature saliency cross-enhancement mechanism is introduced:
\begin{equation}
	\begin{array}{l}
		\bm{\hat {F}}_i^{3 \times 3} = {\rm{softmax}}(\frac{{\bm{\tilde {F}}_{Q,i}^{3 \times 3} \times {{(\bm{\tilde {F}}_{K,i}^{7 \times 7})}^T}}}{{\sqrt d }}) \times \bm{\tilde {F}}_{V,i}^{3 \times 3} + \bm{\tilde {F}}_i^{3 \times 3} \\
		\bm{\hat {F}}_i^{7 \times 7} = {\rm{softmax}}(\frac{{\bm{\tilde {F}}_{Q,i}^{7 \times 7} \times {{(\bm{\tilde {F}}_{K,i}^{3 \times 3})}^T}}}{{\sqrt d }}) \times \bm{\tilde {F}}_{V,i}^{7 \times 7} + \bm{\tilde {F}}_i^{7 \times 7} \\
	\end{array},
\end{equation}
where $\bm T$ denotes the transpose operation. In $\bm{\hat {F}}_i^{3 \times 3}$ and $\bm{\hat {F}}_i^{7 \times 7}$, the larger value, the corresponding focus property is more significant. Therefore, we can use Eq.(11) to synthesize the significant information:
\begin{equation}
	{\bm{\hat {F}}_i}( \cdot , \cdot ) = \left\{ {\begin{array}{*{20}{c}}
			{\bm{\hat {F}}_i^{3 \times 3}( \cdot , \cdot ),if:abs(\bm{\hat {F}}_i^{3 \times 3}( \cdot , \cdot )) \ge abs(\bm{\hat {F}}_i^{7 \times 7}( \cdot , \cdot ))}  \\
			{\bm{\hat {F}}_i^{7 \times 7}( \cdot , \cdot ),if:abs(\bm{\hat {F}}_i^{3 \times 3}( \cdot , \cdot )) < abs(\bm{\hat {F}}_i^{7 \times 7}( \cdot , \cdot ))}  \\
	\end{array}} \right.,
\end{equation}
where $abs( \cdot )$ denotes absolute value operation.

\subsubsection{Hard Pixel Detection}
For focused pixel detection, the features output by FM are sent to an encoder and a decoder to obtain the focus property detection results of each source image, where the encoder consists of five MSFAs and convolutional layers and the decoder consists only five convolutional layers. Thanks to the independent focus detection of each source image, it enables the proposed method to fuse multiple source images in parallel. In this process, the method in this paper needs to separately fuse pixels with ambiguous focus property.

Let $\bm M_{i}$ be focus property detection result of the source image $\bm I_{i}$, and $\bm M_{i}$ is a binary mask comprised of 0 and 1, where 1 indicates that the corresponding pixel in $\bm I_{i}$ is focused, and 0 indicates that the corresponding pixel in $\bm I_{i}$ is defocused. Since each source image has an independent detection result, the hard pixels can be identified by finding inconsistencies between detection results. For $N$ source images, $\bm M_{1}(x,y)+\bm M_{2}(x,y)+\cdots+\bm M_{N}(x,y)=1$  means that the focus property of the pixel at $(x,y)$ can be determined, otherwise the pixel at $(x,y)$ is identified regarded as hard pixel. The process can be formulated as:
\begin{equation}
\begin{aligned}
    \bm {M}_h(x,y) = \left\{ \begin{array}{l}
    	0,if:\bm {M}_1(x,y) + \bm {M}_2(x,y) +  \cdots  + \bm {M}_N(x,y) = 1 \\
    	1/2,{\rm{  otherwise}} \\
    \end{array}\right.,
\end{aligned}
\end{equation}
where $\bm M_{h}(x,y) = 0$ indicates that the focus property of pixel at location $(x,y)$ is determined, $\bm M_{h}(x,y) = 1/2$ indicates that the pixel at location $(x,y)$ is hard pixel.
\subsection{Full-focus Image Generation With Multi-directional Feature Embedding}
\begin{figure}[t!]
	\centering
	\includegraphics[width=2.7in,height=1.4in]{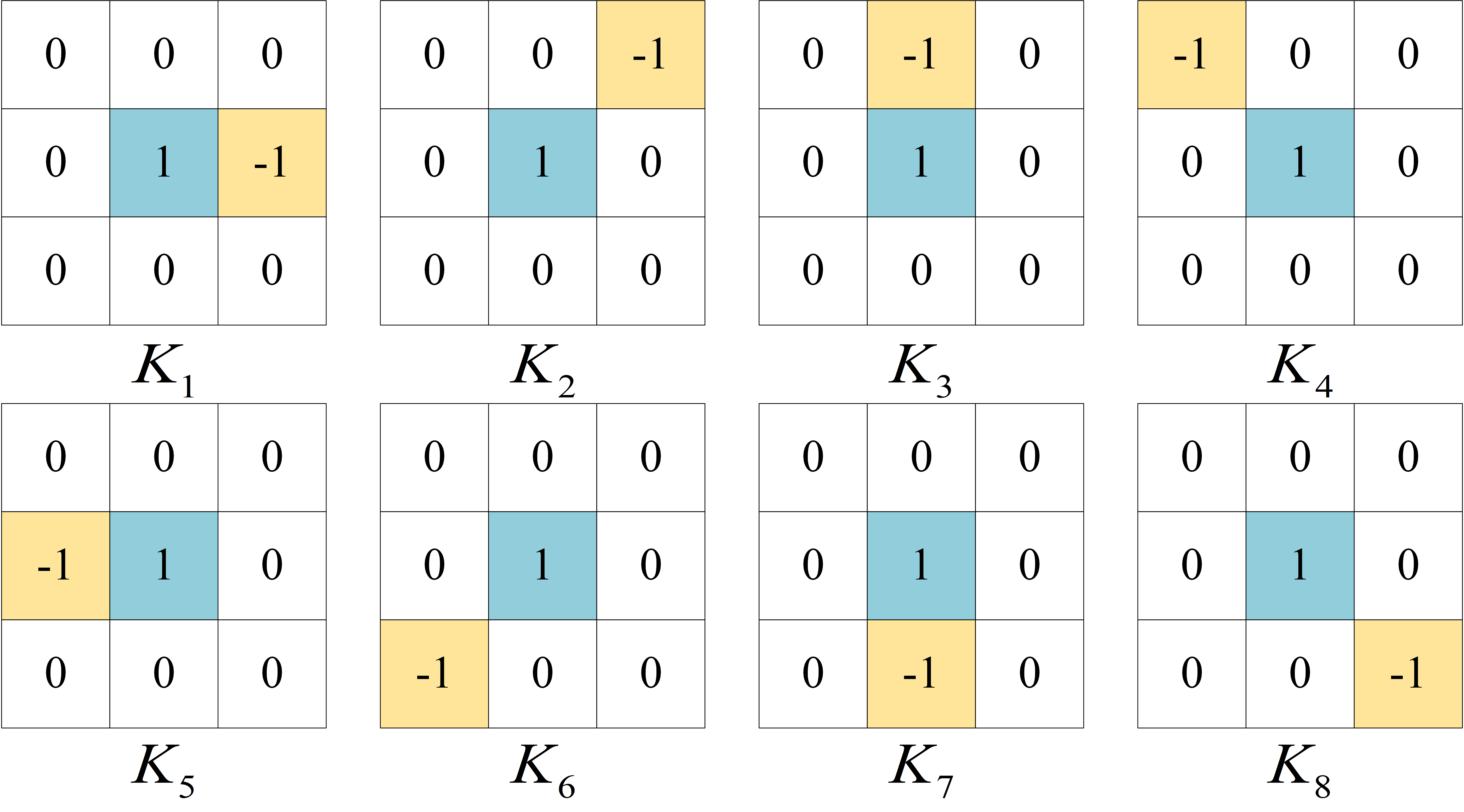}
	\setlength{\abovecaptionskip}{0.cm}
	\caption{Illustration of eight directional edge extraction kernels.}
	\vspace{-0.4cm}
	\label{labe5}
\end{figure}
Due to the ambiguity of the focus properties of hard pixels, it's not optimal to construct the final fused result by selecting the focused pixels from source images. To reduce the negative effect of hard pixels on fusion result, we devise a full-focus image generation method with multi-directional feature embedding (FFIG-MFE) to construct the fusion result of hard pixels. FFIG-MFE consists of multi-directional edge extraction (MDEE), weight generator and edge feature embedding (EFE). As shown in Fig.\ref{labe2}, multi-directional edge extraction in FFIG-MFE can be achieved by the designed multi-directional edge extraction kernels presented in Fig.\ref{labe5}. Specifically, the size of each kernel is ${3 \times 3}$ with the value of center position being 1. In FFIG-MFE, there are eight kernels corresponding to eight different directions are used to extract the edges from the source images. The multi-directional edges are denoted as:
 \begin{equation}
	\begin{aligned}
		\bm {F_{i,{k_n}}} = abs(\bm I_{i}\circledast \bm K_{n})~~(n = 1,2,3...8)
	\end{aligned},
\end{equation}
where $\bm F_{i,{k_n}}$ denotes the edge features of the source image $\bm I_{i}$ in the $n$-th direction.
\begin{figure}[t!]
	\centering
	\includegraphics[width=3.4in,height=1.6in]{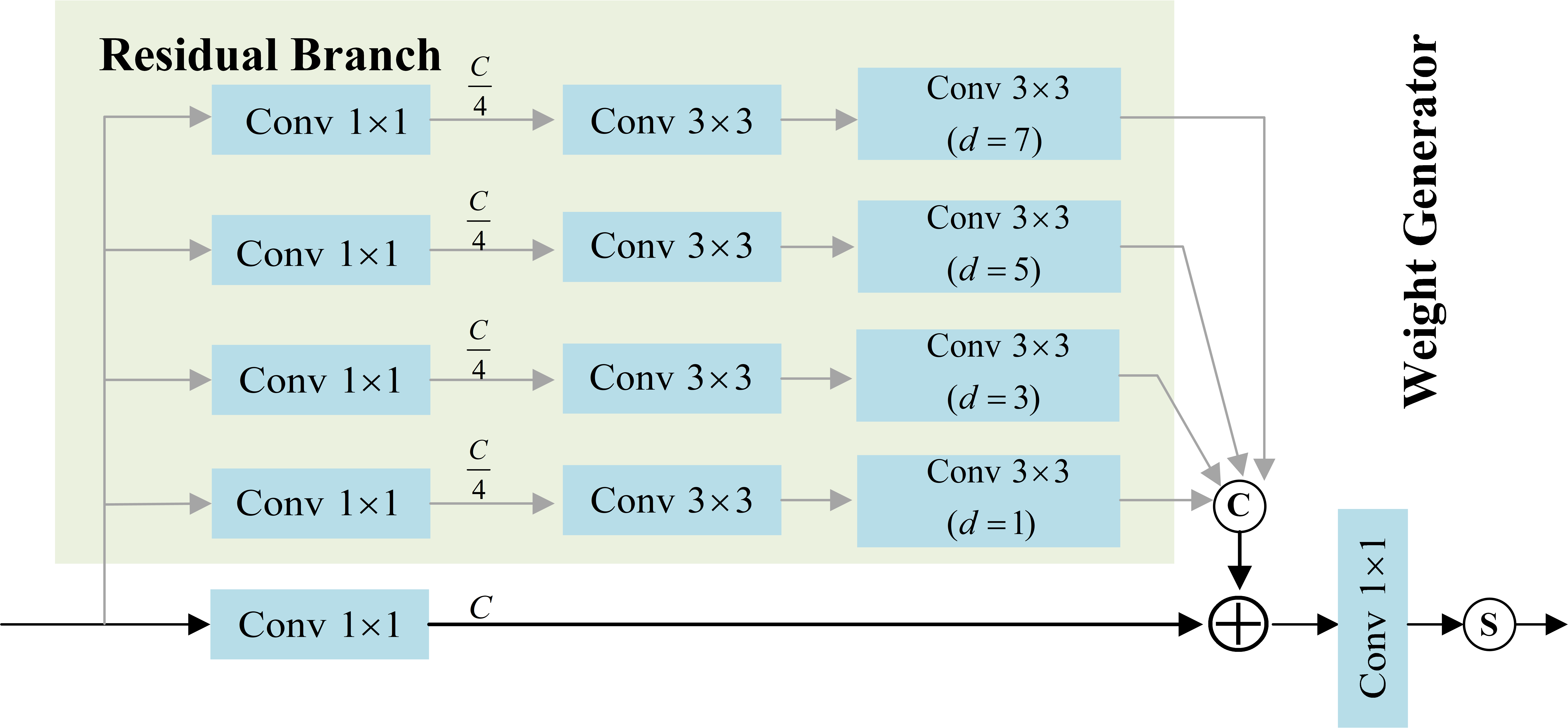}
	\caption{{The detailed architecture of the weight generator, where the residual branch is used to extract features under different receptive fields. More appropriate weights are generated by integrating features from multiple scales.}}
	\vspace{-0.4cm}
	\label{labe6}
\end{figure}

\begin{figure}[!t]
	\centering
	\includegraphics[width=3.4in,height=1.8in]{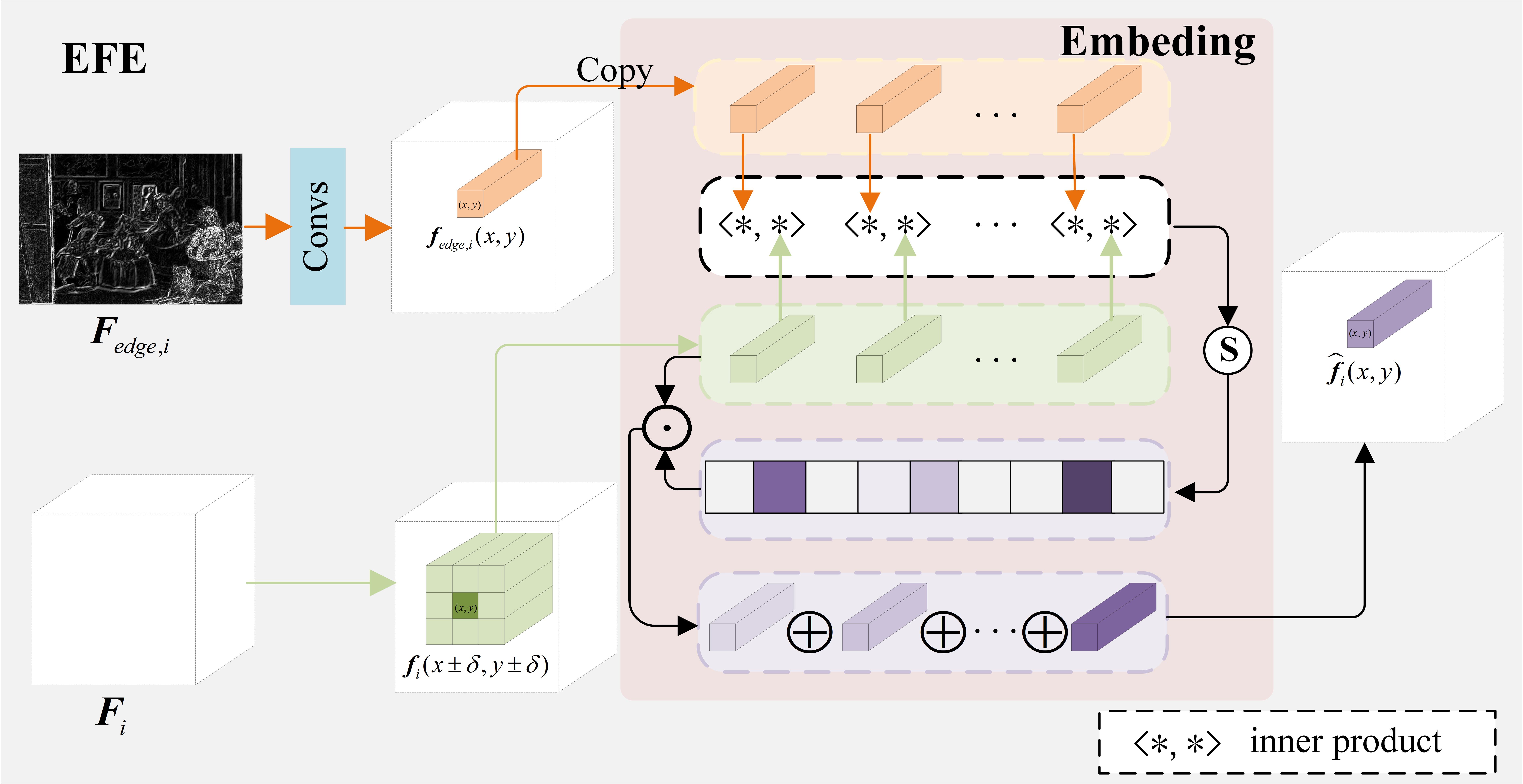}
	\setlength{\abovecaptionskip}{0.cm}
	\caption{The detailed architecture of EFE.}
	\vspace{-0.4cm}
	\label{labe7}
\end{figure}

To fully utilize the features at eight direction to generate a full-focus image, a weight generation mechanism is designed to dynamically produce the weights for integrating multi-direction features. As shown in Fig.\ref{labe6}, the weight generator consists of ${1 \times 1}$ convolutional layer and a residual branch, which first employs, ${1 \times 1}$ convolutional operation  to reduce the number of channels to $1/4$ of the original number, and then utilizes a ${3 \times 3}$ convolutional operation and a ${3 \times 3}$ convolutional operation with dilation rate of $d (d = 1,3,5,7)$ to extract features under different respective field. The outputs of ${1 \times 1}$ convolutional layer and residual block are concatenated, and then weights $\omega _i^n(x,y)$ can be achieved after ${1 \times 1}$ convolution and softmax. With $\omega _i^n(x,y)$, the integrated multi-directional edge features can be represented as:
 \begin{equation}
	\begin{aligned}
		{\bm {F}}_{edge,i}^{}(x,y) = \sum\limits_{n = 1}^8 {\omega _i^n} (x,y) \odot \bm F_{i,{k_n}}^{}(x,y)
	\end{aligned}.
\end{equation}
Since $\bm {F}_{edge,i}$ contains the focus property of source image $\bm I_{i}$, we use it to highlight the effect of focused features in EFE for generating a full-focus image. As shown in Fig.\ref{labe7}, we extract the features from $\bm {F}_{edge,i}$ using ${1 \times 1}$ convolution to make the dimension consistent with $\bm {F}_{i}$. Let $\bm f_{edge,i}(x,y)$ be the feature vector composed of elements from different channels of $\bm {F}_{edge,i} $ at location $(x,y)$. The result is represented as $\bm {F}_i$ after $\bm I_{i}$ is processed by convolutional operation, BN and ReLU activate function. We measure the similarity between $\bm f_{edge,i}^{}(x,y)$ and all feature vectors $\bm f_{i}(x \pm \delta, y \pm \delta )$ by inner product, where $\bm f_{i}(x \pm \delta, y \pm \delta )$ denotes the feature vector within $\left[ {x \pm \delta ,y \pm \delta } \right]$. Then, the weight is obtained after softmax activated:
\begin{equation}
	\begin{aligned}
		\bm D_{i}(x',y') = softmax(\bm f_{edge,i}^{}(x,y)\bm f_i^T(x',y'))
	\end{aligned},
\end{equation}
where $(x', y') \in \left[ {x \pm \delta ,y \pm \delta } \right]$. The weight is used to scale all feature vectors $\bm f_{i}(x \pm \delta, y \pm \delta )$:
\begin{equation}
	\begin{aligned}
		{{{\hat{\bm f}}}_i}(x,y) = \sum\limits_{y' = y - \delta }^{y + \delta } {\sum\limits_{x' = x - \delta }^{x + \delta } {\bm D_{i}(x',y')\bm f_i^{}(x',y')} }
	\end{aligned}.
\end{equation}
When $(x,y)$ in ${\hat{\bm f}}_{i}(x,y)$ traverses all pixels of the entire image, the features for generating a full-focus image are generated.
Let
\begin{equation}
	\begin{aligned}
		\hat{\bm F}(x,y)=\max\{{\hat{\bm f}}_{1}(x,y)+{\bm f}_{1}(x,y), \cdots, {\hat{\bm f}}_{N}(x,y)+{\bm f}_{N}(x,y)\}
	\end{aligned}.
\end{equation}
Since $\bm M_{h}$ can represent the difficulty of focused pixel detection, we concatenate it with $\hat{\bm F}$ and then feed it to the decoder to generate full-focus image $\bm F_{g}$:
 \begin{equation}
	\begin{aligned}
		\bm F_{g} = \bm E_{d}([2\bm M_{h},\hat{\bm F}])
	\end{aligned},
\end{equation}
where $\bm E_{d}$ denotes a full-focus image decoder consisting of a residual block and a convolutional layer.

\subsection{Fusion Result Construction Guided by Hard Pixels}
With $\bm M_{h}$ and full-focus image $\bm F_{g}$, a hard pixels guided fusion result construction is proposed. If the pixel does not belong to the hard pixels, the fused result is constructed directly by selecting the corresponding pixel from source images. Otherwise, the corresponding pixel of $\bm F_{g}$ is selected. To this end, the Eq.(19) is developed to update the decision map for fusion result construction:
\begin{equation}
	\begin{aligned}
		{\tilde {\bm M}}_{i}(x,y) = \left\{ \begin{array}{l}
			1,{if:}\bm M_{i}(x,y) + \bm M_{h}(x,y) = 1 \\
			0,{\rm{otherwise}} \\
		\end{array} \right.
	\end{aligned},
\end{equation}
where $	{\tilde {\bm M}}_{i}(x,y) = 1$ indicates that the pixel located at $(x,y)$ is the focused pixel in the source image $\bm I_{i}$. The final fused result guided by ${\bm{M}}_{h}$ can be expressed as:
\begin{equation}
	\begin{aligned}
		\bm F = {\bm{\tilde {M}}_1} \odot \bm I_{1} + {\bm{\tilde {M}}_2} \odot \bm I_{2} +  \cdots  + {\bm{\tilde {M}}_N} \odot \bm I_{N} + 2\bm M_{h} \odot \bm F_{g}
	\end{aligned}.
\end{equation}

\subsection{Loss Function}
The loss functions of the method in this paper are divided into two parts: the hard pixels detection loss and the full-focus image generation loss:

\textbf{Hard pixels detection loss:} In HPD, cross entropy loss formulated in Eq.(21) is used to enable the network to realize the detection of pixel focus property:
\begin{equation}
\begin{aligned}
{\ell _{ce}} =  - \sum\limits_{i = 1}^N {\sum\limits_{x = 1,y = 1}^{H,W} {({g_i}(x,y)\log ({p_i}(x,y)) + } } \\
 (1 - {g_i}(x,y)\log (1 - {p_i}(x,y))
\end{aligned},
\end{equation}
where ${p_i}(x,y)$ denotes the focused probability at position $(x,y)$ for the $i$th source image, and ${g_i}(x,y)$ is its corresponding label.

\textbf{Full-focus image generation loss:} To obtain a clear full-focus image, besides the global reconstruction loss, the local reconstruction loss is also used in the hard pixel region:
\begin{equation}
	\begin{aligned}
		{\ell _{rec}} = {\left\|\bm F_{g} - \bm G_{f} \right\|_1} + \lambda {\left\| {\bm M_{h} \odot \bm F_{g} - \bm M_{h} \odot \bm G_{f}} \right\|_1}
	\end{aligned},
\end{equation}
where $\bm G_{f}$ denotes the label of generated full-focus image, $\lambda$ is a hyperparameter that balances the effects of global and local loss.
\section{Experiments}\label{section4}
\subsection{Datasets}
In training of GRFusion, we construct training set by artificial synthesis. Similar to DRPL\cite{26}, 200 all-in-focus images selected from ImageNet Large Scale Visual Recognition Challenge 2012 (ILSVRC2012)\cite{38} are used to synthesize the training samples. To achieve the data augmentation, each raw image is randomly cropped to 9 sub-images whose sizes are ${256 \times 256}$. Let a ${256 \times 256}$ sub-image be $\bm I$, a pair of multifocus images is synthesized by:
\begin{equation}
	\begin{aligned}
		\begin{array}{l}
			\bm I_{1} = \bm M \odot \bm I + (\bm 1 - \bm  M) \odot {\bm \tilde{G}}(\bm I) \\
			\bm I_{2} = \bm M \odot {\bm \tilde{G}}(\bm I) + (\bm 1 - \bm M) \odot \bm I \\
		\end{array},
	\end{aligned}
\end{equation}
where $\bm M$ denotes a binary mask randomly generated by the ``findContours" function in OpenCV, and ${\bm \tilde{G}}( \cdot )$ is Gaussian filter with five different blurred versions. In the testing, the datasets used in our experiments are Lytro\cite{45}, MFI-WHU\cite{28} and MFFW\cite{39}. The details of the testing datasets are shown in Table \ref{table1} and the source images used to show the fused results are displayed in Fig.\ref{labe8}:
\begin{figure}[!t]
	\centering
	\includegraphics[width=3.4in,height=2in]{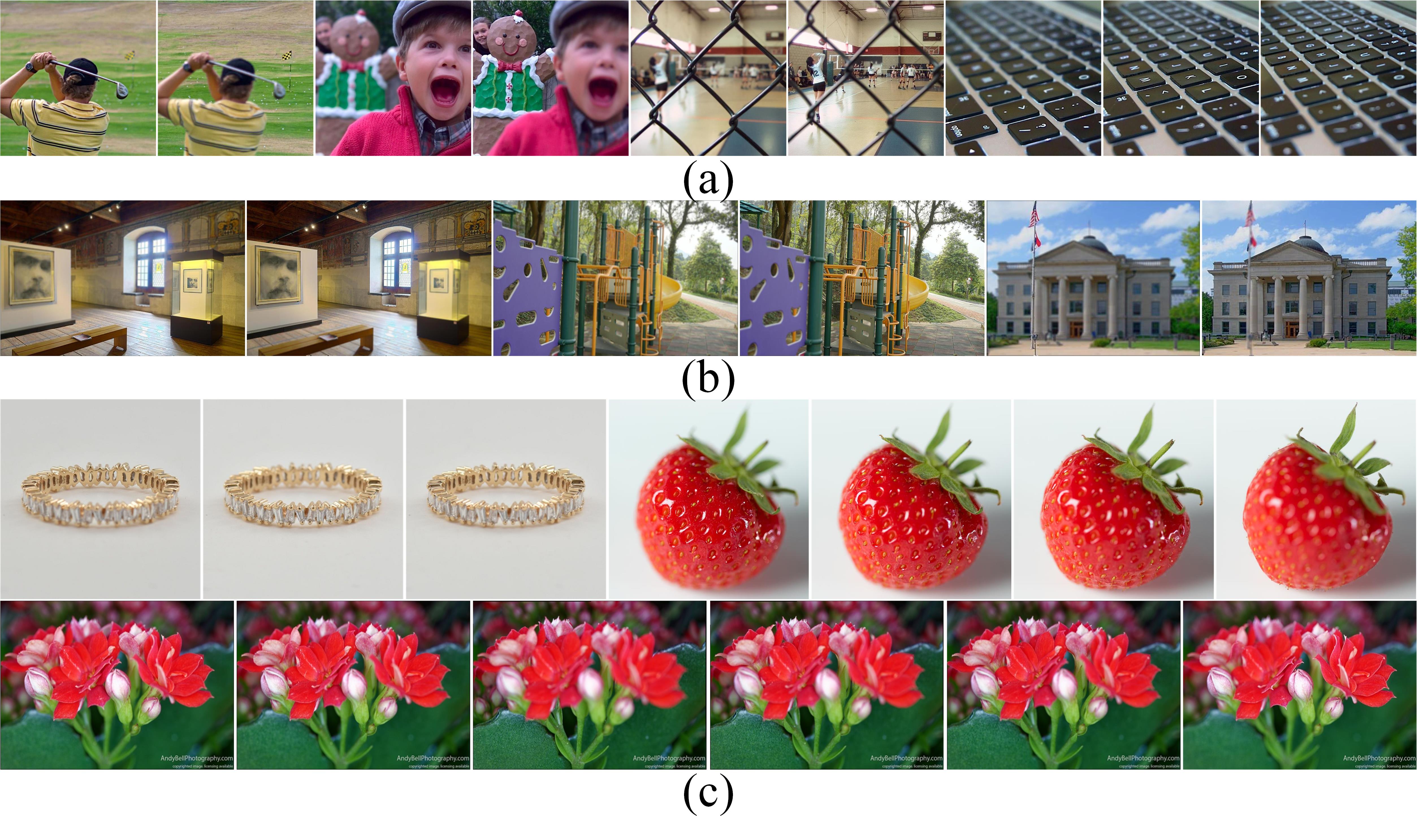}
	\setlength{\abovecaptionskip}{0.cm}
	\caption{Source images used for visual comparison of the fused results. (a) Source images from ``Lytro". (b) Source images from ``MFI-WHU". (c) Source images from ``MFFW".}
	\label{labe8}
\end{figure}
\begin{table}[!t]\footnotesize
	\centering {
		\caption{The details of testing datasets.}
		\label{table1}
		\renewcommand\arraystretch{1.0}
		\begin{tabular}{c c c c}
			\hline
			Dataset & Resolution & No.of images & Size \\
			\hline
			
			\multirow{2}*{Lytro} & \multirow{2}*{${520 \times520}$} & 2  & 20  \\
			
			&  & 3 & 4 \\
			\hline
			
			MFI-WHU & Various & 2 & 120 \\
			\hline
			
			\multirow{4}*{MFFW} & \multirow{4}*{Various} & 2 & 13 \\
			&  & 3 & 4 \\
			&  & 4 & 1 \\
			&  & 6 & 1 \\
			\hline
	\end{tabular}}
\end{table}
\subsection{Implementation Details}
The training process includes two stages: the training of HPD and the training of FFIG-MF. FFIG-MF is trained after the training of HPD.  In training of HPD, the total epoch is set to 600 with batch size of 24. In training of FFIG-MF, the total epoch is also set to 600 and the batch size is set to 4. The loss functions in Eq.(21) and Eq.(22) are minimized with the Adam Optimizer \cite{40}. The learning rate is initialized to 0.0001 and the exponential decay rate is set to 0.9.  The proposed method is implemented on Pytorch framework and the experiments are conducted on a  desktop with a single GeForce RTX 3090.

\begin{figure*}[t!]
	\centering
	\includegraphics[width=6in,height=8.9in]{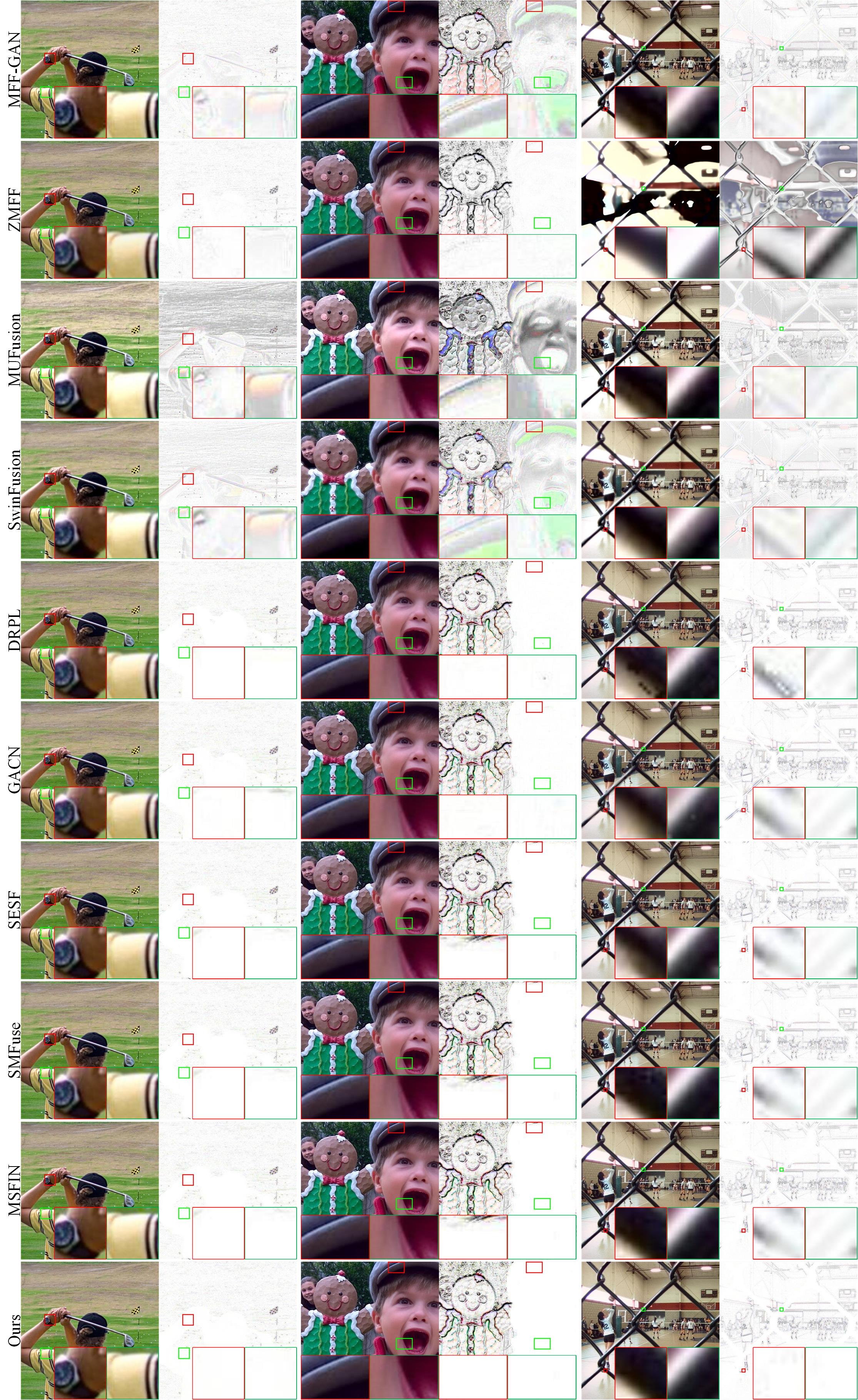}
	\setlength{\abovecaptionskip}{0.cm}
	\caption{Fused results of different methods on ``lytro-01", ``lytro-03" and ``lytro-05" image pairs. For each image pair, the results in the left represent the fused images and the right represent the differences  between the fused image and the corresponding one source image. For the boxed areas, the less remaining information indicates better fusion results.}
	\vspace{-0.4cm}
	\label{labe9}
\end{figure*}
\begin{figure*}[t!]
	\centering
	\includegraphics[width=7in,height=8.2in]{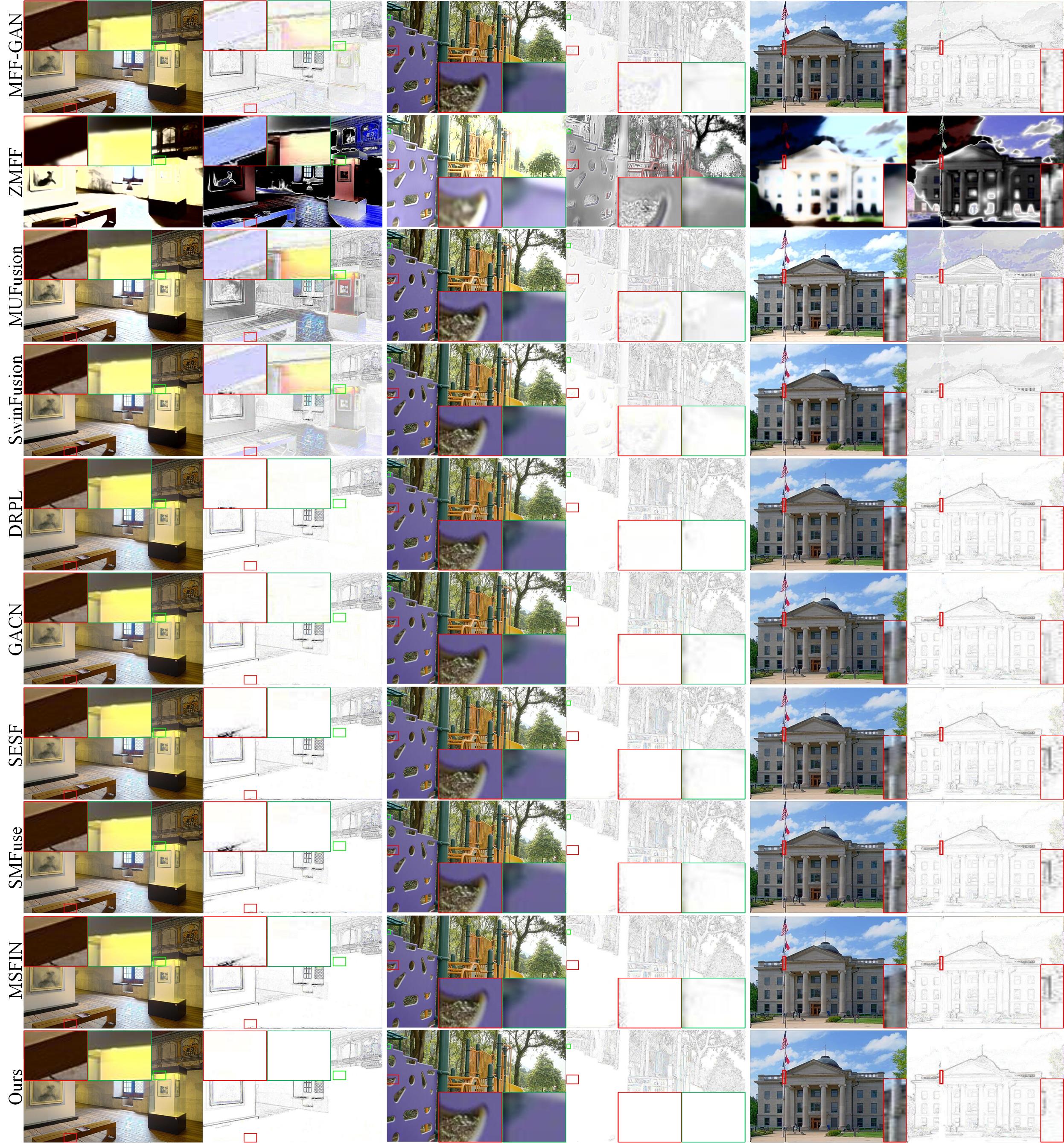}
	\setlength{\abovecaptionskip}{0.cm}
	\caption{Fused results of different methods on ``MFI-WHU-11", ``MFI-WHU-17" and ``MFI-WHU-05" image pairs. For each image pair, the results in the left represent the fused images and the right represent the differences  between the fused image and the corresponding one source image. For the boxed areas, the less remaining information indicates better fusion results.}
	\vspace{-0.4cm}
	\label{labe10}
\end{figure*}
\begin{figure*}[t!]
	\centering
	\includegraphics[width=7in,height=7.4in]{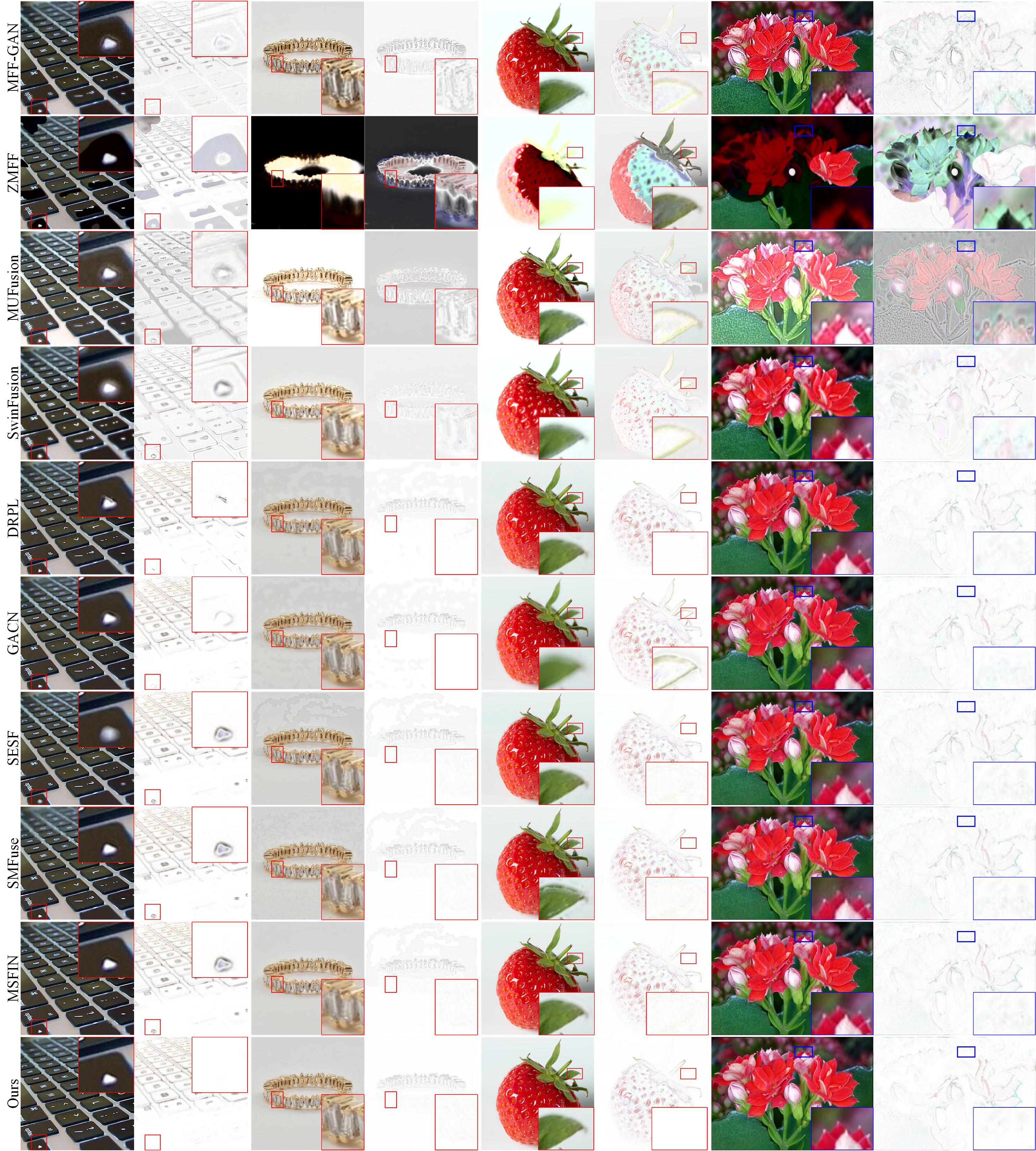}
	\setlength{\abovecaptionskip}{0.cm}
	\caption{Fused results with multiple source images. The fusion results of ``Lytro-triple series-03", ``MFFW3-03", ``MFFW3-06" and ``MFFW3-02" are presented in the first, third, fifth and seventh columns.}
	\vspace{-0.4cm}
	\label{labe11}
\end{figure*}

\begin{table*}[!t]\footnotesize
	\centering {
		\caption{Quantitative comparison of fusion results on ``Lytro" and  ``MFI-WHU" datasets. The red font represents the optimal result, while the blue font represents the suboptimal result.}
		\label{table2}
		\renewcommand\arraystretch{1.0}
		\begin{tabular}{c c c c c c c c c c c c c }
			\hline
			\multirow{2}*{Methods} & \multicolumn{5}{c }{Lytro dataset} & \multicolumn{5}{c }{MFI-WHU dataset} \\
			\cline{2-11}
			&${Q_{MI}}$ &${Q_{AB/F}}$&${Q_{CB}}$&${Q_{NCIE}}$&${Q_{SSIM}}$&${Q_{MI}}$ &${Q_{AB/F}}$&${Q_{CB}}$&${Q_{NCIE}}$&${Q_{SSIM}}$\\
			\hline
			
			MFF-GAN\cite{28}
			&0.80995&0.64471&0.61022&0.82540&1.65349&0.76162&0.62241&0.60932 &0.82352&1.64444\\
			
			ZMFF\cite{17}
			&0.90869&0.67569&0.69665&0.83001&1.72113&0.71858&0.61924&0.66719&0.82227&1.63110\\
			
			MUFusion\cite{29} &0.73928&0.61593&0.59090&0.82231&1.60150&0.65023&0.56568&0.56337&0.81991&1.56340\\
			
			SwinFusion\cite{32} &0.81836&0.66624&0.61071&0.82555&1.66042&0.76030&0.64815&0.63719&0.82306&1.64592\\
			
			DRPL\cite{26} &1.09872&0.73257&0.76223&0.84132&1.72350&1.07076&0.71212&0.78625&0.83999&1.71368 \\
			
			GACN\cite{27} &1.12565&0.73649&0.77273&0.84336&1.72347&1.09467&0.70672&0.79610&0.84212&1.70880\\
			
			SESF\cite{20} &1.17177&0.74152&0.79114&0.84681&\textcolor{blue}{\bf1.72534}&\textcolor{blue}{\bf1.20181}
			&0.72097 &0.80893& 0.84908&1.71498\\
			
			SMFuse\cite{22}
			&1.16521&0.74007&0.77957&0.84628&1.72435&1.17048&\textcolor{blue}{\bf0.72933}&0.81081&0.84724&1.71463\\
			
			MSFIN\cite{23}
			&\textcolor{blue}{\bf1.18912}&\textcolor{blue}{\bf0.74739}&\textcolor{red}{\bf0.79569}
			&\textcolor{blue}{\bf0.84747}&\textcolor{red}{\bf1.72547}&1.19948&0.72820
			&\textcolor{blue}{\bf0.82107}&\textcolor{blue}{\bf0.84911}&\textcolor{blue}{\bf1.71533}\\
			
			GRFusion
			&\textcolor{red}{\bf1.18998}&\textcolor{red}{\bf0.74742}&\textcolor{blue}{\bf0.79497}
			&\textcolor{red}{\bf0.84798}&1.72495&\textcolor{red}{\bf1.20633}&\textcolor{red}{\bf0.73382}&\textcolor{red}{\bf0.82473}&\textcolor{red}{\bf0.85014}&\textcolor{red}{\bf1.71536}\\
			\hline
	\end{tabular}}
\end{table*}

\subsection{Comparison Methods and Evaluation Metrics}
To demonstrate the effectiveness of our GRFusion, both qualitative and quantitative quality assessment are performed. In this process, we compare GRFusion with 9 state-of-the-art methods. These methods can be divided into two categories, \textit{i.e.},  feature reconstruction-based methods including MFF-GAN\cite{28}, ZMFF\cite{17}, MUFusion\cite{29} and SwinFusion\cite{32}, pixel recombination-based methods including  DRPL\cite{26}, GACN\cite{27}, SESF\cite{20}, SMFuse\cite{22} and MSFIN\cite{23}. To quantitatively evaluate the fusion results, five commonly used metrics including mutual information(${Q_{MI}}$\cite{41}), edge based similarity metric(${Q_{AB/F}}$\cite{43}), chen-blum metric(${Q_{CB}}$\cite{44}), nonlinear correlation information entropy(${Q_{NCIE}}$\cite{42}) and structural similarity (${Q_{SSIM}}$\cite{57}) are used in our experiments. In these metrics, ${Q_{MI}}$  evaluates how much information of the source images is included in the fusion result. ${Q_{AB/F}}$ measures the degree of edges of source images transferred into the fused result.  ${Q_{CB}}$ assesses the amount of information transferred from source images into fused result. ${Q_{NCIE}}$ evaluates the fused image from preservation of detail, structure, color and brightness. ${Q_{SSIM}}$  measures the structural similarity between the fusion result and the source images.  The larger the values of these evaluation indicators, the better the quality of the fusion results.

\subsection{Fusion of Two Source Images}
The first experiment is performed on fusion of two source images. In this process, the image pairs from ``Lytro" and ``MFI-WHU" are used as the test samples, and the fusion results of the source image pairs shown in Fig.\ref{labe8} are displayed in Fig.\ref{labe9} and Fig.\ref{labe10}. As observed, GRFusion performs observable advantages when compared to the other 9 methods. First of all, GRFusion is deigned by combining feature reconstruction-based method and focused pixel recombination-based method. Therefore, it can maximize the retention of focused information in the source image, avoiding the loss of detailed information and the introduction of artifacts.
Meanwhile, thanks to FRC-G-HP, the degradation of fusion results caused by focus detection errors can be effectively alleviated. Consequently, the developed method can produce the results with better visual quality as shown in Fig.\ref{labe9} and Fig.\ref{labe10}.
\begin{figure*}[!t] \centering
	\subfigure[Values of $Q_{MI}$]  {\includegraphics[height=1.0in,width=1.3in,angle=0]{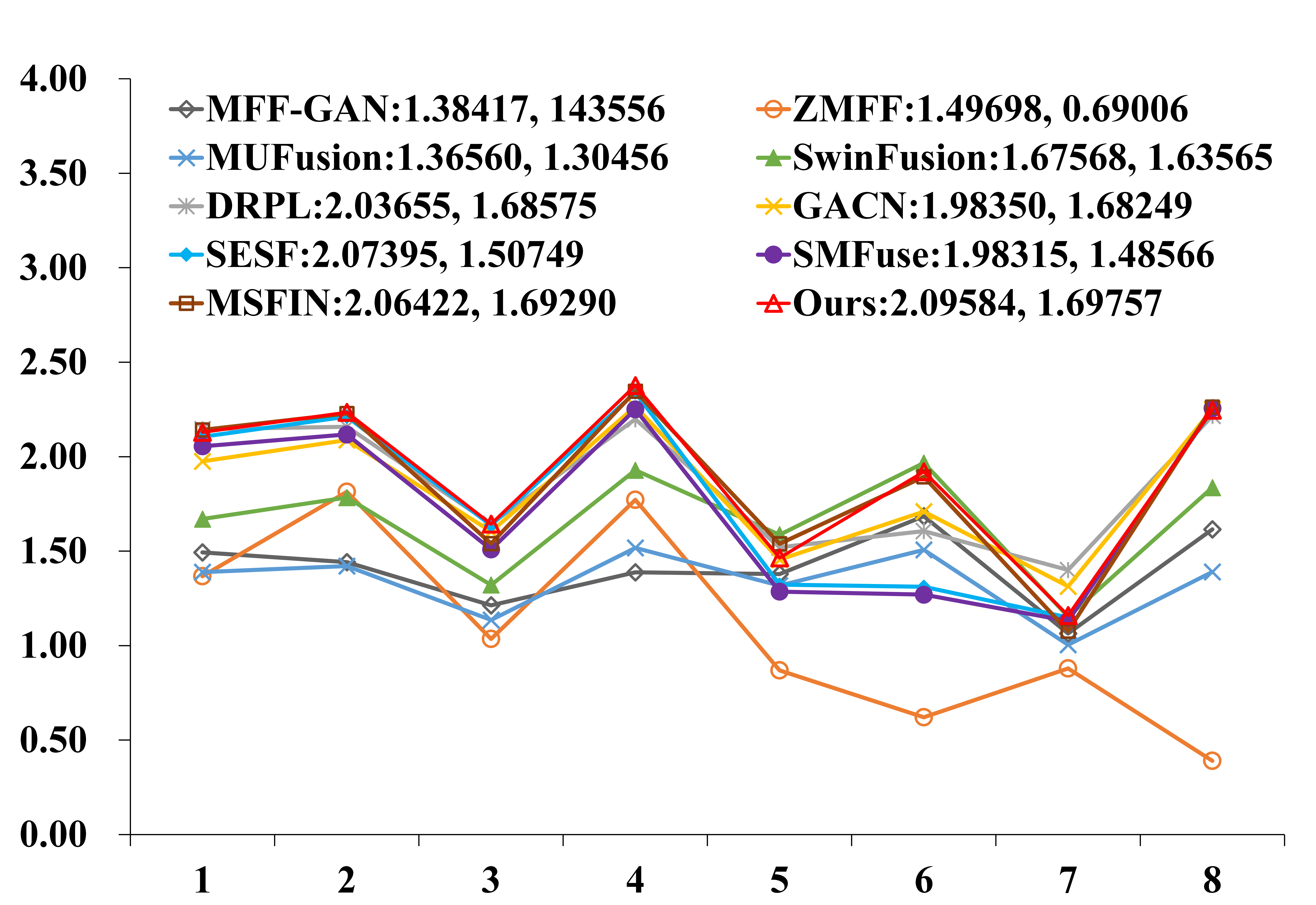}}
	\subfigure[Values of $Q_{AB/F}$]   {\includegraphics[height=1.0in,width=1.3in,angle=0]{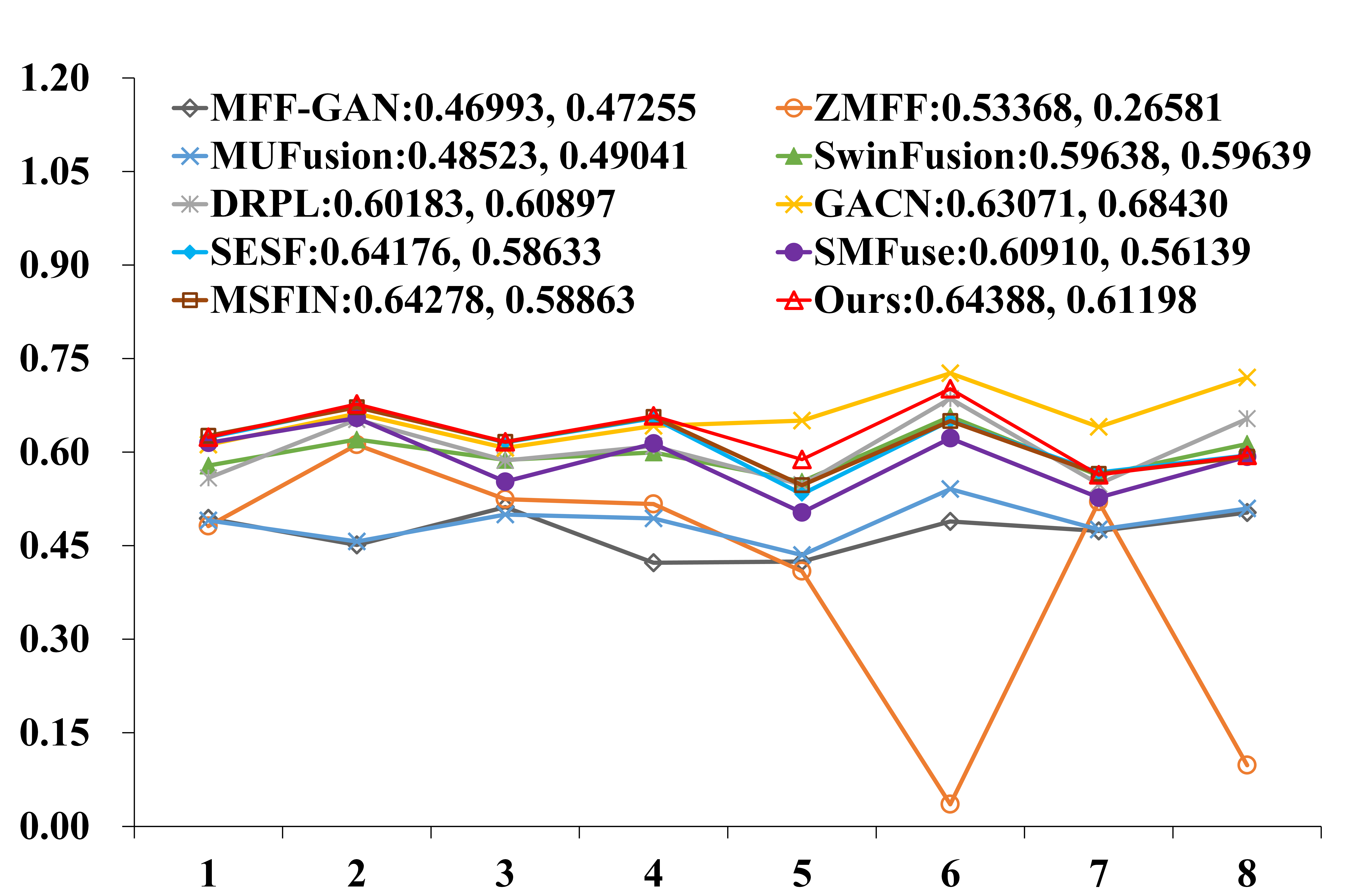}}
	\subfigure[Values of $Q_{CB}$]  {\includegraphics[height=1.0in,width=1.3in,angle=0]{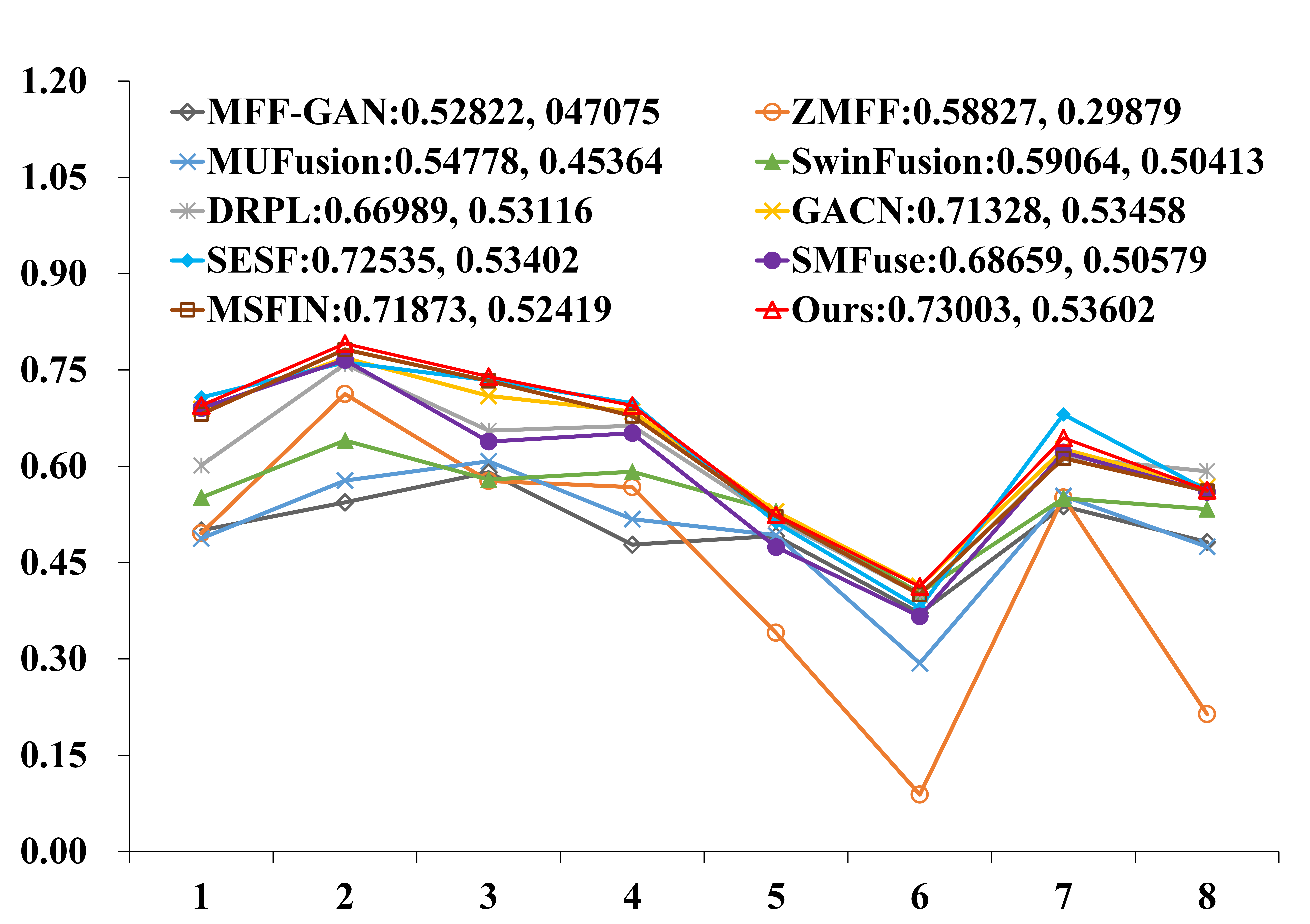}}
	\subfigure[Values of $Q_{NCIE}$]   {\includegraphics[height=1.0in,width=1.3in,angle=0]{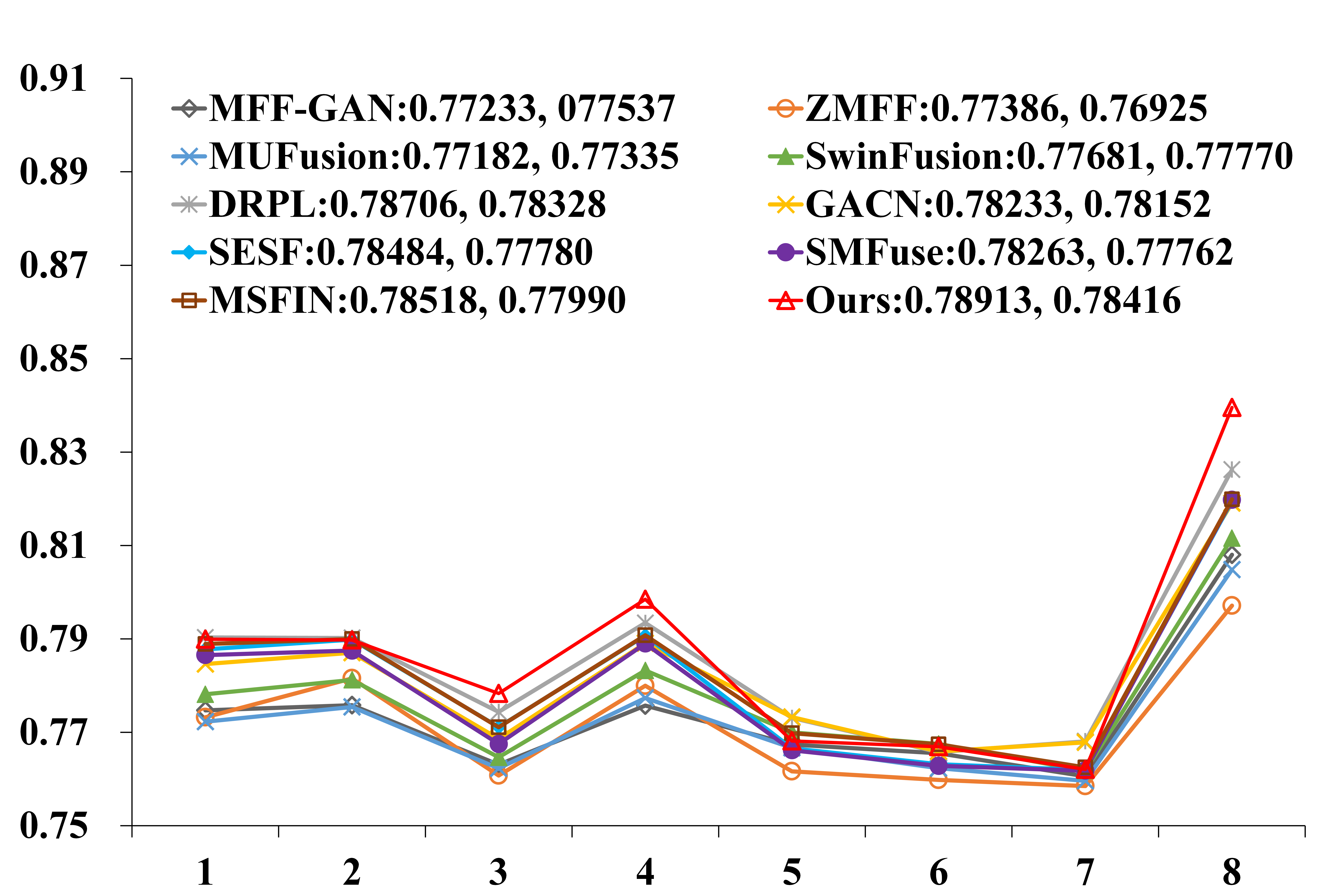}}
	\subfigure[Values of $Q_{SSIM}$]   {\includegraphics[height=1.0in,width=1.3in,angle=0]{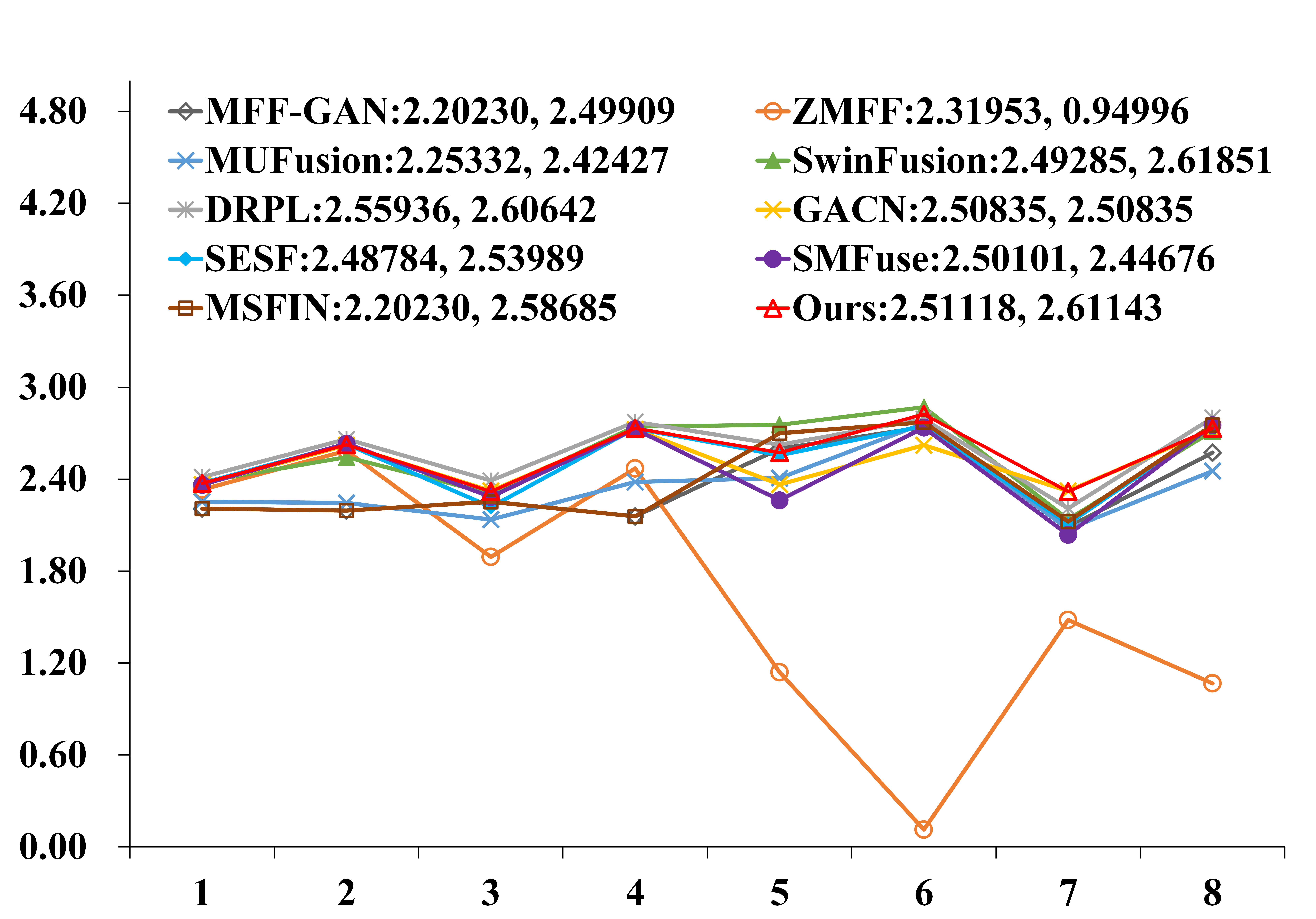}}
	\caption{Objective evaluation of fusion results of multiple source images.  The horizontal axis represents the source image group label, where 1--4 represents the first 4 source image groups from Lytro, and 5-8 represents the second 4 source image groups from MFFW.}
	\vspace{-0.4cm}
	\label{labe12}
\end{figure*}

Specifically, from the boxed area we can see that the residual information in the results generated by the proposed method is less than that of the compared methods. This indicates that GRFusion can transfer more focused information of the source images into the fusion results. Moreover, we can also find that MFF-GAN, MUFusion and SwinFusion methods suffer from chromatic aberrations, and ZMFF tends to loss the information in focused regions. The results obtained by the focused pixel recombination-based methods, \textit{i.e.}, DRPL, GACN, SESF, SMFuse and MSFIN are prone to blurring at the boundary of the focus region. This can be seen from the residual information in the boxed areas. To evaluate the fused results of each method more objectively, we further list quantitative comparison results in Table \ref{table2}. As we can see, the results generated by the proposed method are optimal in most matrices. It further demonstrates the effectiveness of the proposed method.

\subsection{Fusion of Multiple Source Images}
Unlike most of the existing methods, GRFusion is able to achieve the parallel fusion of multiple images and shows superior fusion performance. To validate this argument, the sequence sample set in Lytro and MFFW with more than two  multifocus source images are tested. Specifically, there are 3, 4 and 6 source images in `Lytro-triple series-03", ``MFFW3-03", ``MFFW3-06" and ``MFFW3-02". The fusion results generated by different methods are shown in Fig. \ref{labe11}. From the boxed areas shown in the first and the second columns, it can be seen that GRFusion can not only achieve fusion processing of three source images, but also more effectively preserve the focus information, resulting in a fusion image with higher visual quality. From the remaining results illustrated in Fig. \ref{labe11}, we can draw similar conclusions. This is because the method proposed in this paper can perform parallel fusion processing on multiple source images, avoiding the introduction of multiple feature extraction used in alternating fusion. Therefore, the proposed method can effectively alleviate the information loss caused by insufficient feature extraction. To objectively evaluate the quality of fusion results, the evaluation results of five metrics on the fusion images shown in Fig. \ref{labe11} are displayed in Fig. \ref{labe12}. The objective evaluation results further prove that the proposed method also has advantages when fusing more than two source images.
\subsection{Ablation Study}
\subsubsection{Ablation Study on Network Structure}
\begin{figure}[!t]
	\centering
	\includegraphics[width=3.2in,height=1.4in]{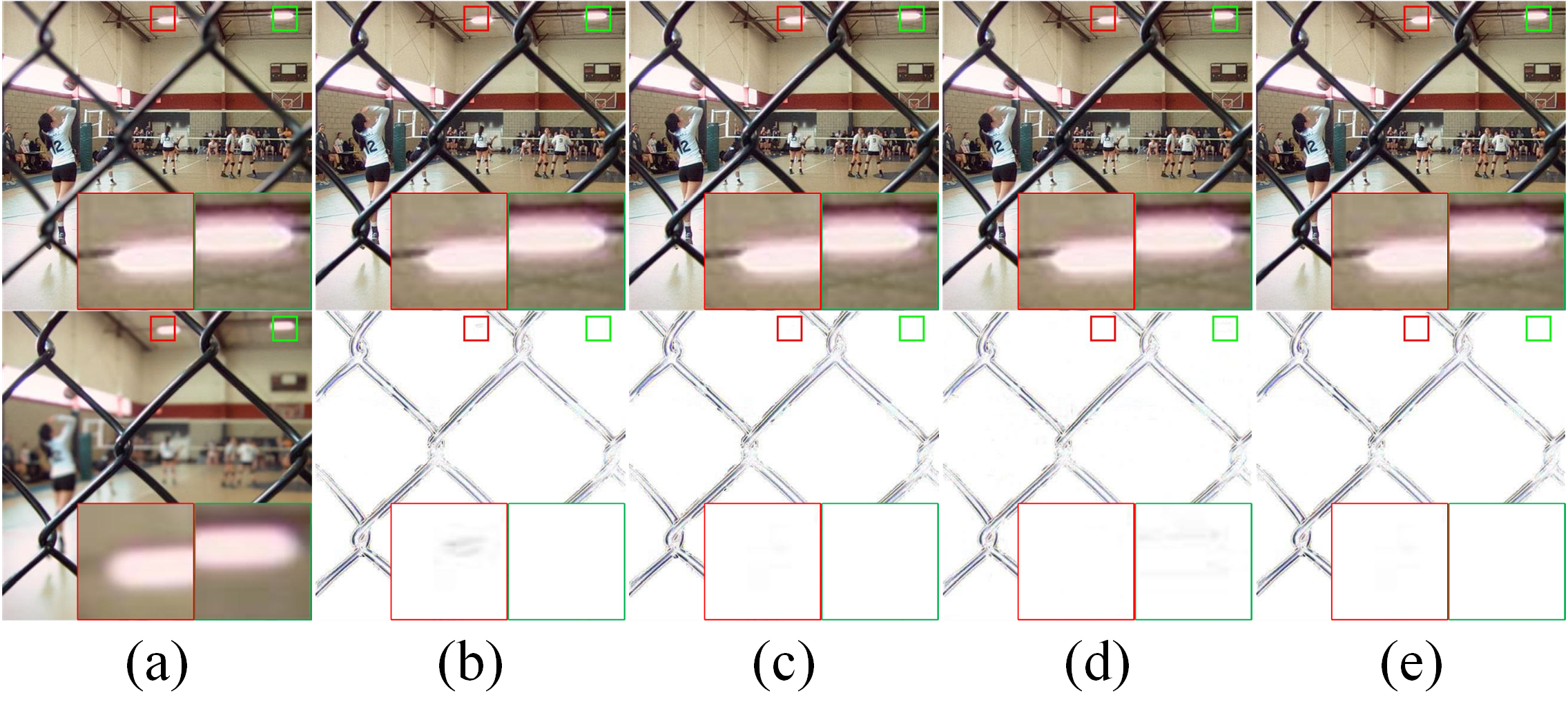}
	\setlength{\abovecaptionskip}{0.cm}
	\caption{Ablation study on network structure. (a) Source images (b) mode1 (c) mode2 (d) mode3 (e) GRFusion.}
	\vspace{-0.4cm}
	\label{labe14}
\end{figure}

\begin{table}
	\caption{Ablation study on network structure.}
	\label{table3}
	\resizebox{1.0\linewidth}{!}{
		\begin{tabular}{c c c|c c c c c c}
		\hline
		mode1&mode2&mode3&${Q_{MI}}$&${Q_{AB/F}}$&${Q_{CB}}$&${Q_{NCIE}}$&${Q_{SSIM}}$\\
		\hline
		\checkmark & & & 1.18157 & 0.74660 & \textcolor{blue}{\bf 0.79488} & 0.84702& 1.72472	\\
		
		  &\checkmark & & \textcolor{blue}{\bf 1.18402} & \textcolor{blue}{\bf 0.74698} & 0.79441 & \textcolor{blue}{\bf 0.84765}&\textcolor{blue}{\bf1.72493}\\
		
		  & &\checkmark & 1.11146 & 0.73550 & 0.75651 & 0.84253&1.72491\\
		
		 \checkmark& \checkmark &\checkmark & \textcolor{red}{\bf 1.18998} & \textcolor{red}{\bf 0.74742} & \textcolor{red}{\bf 0.79497} & \textcolor{red}{\bf 0.84798}&\textcolor{red}{\bf1.72495}\\
		\hline
		\end{tabular}
	}
\end{table}
To verify the effectiveness of combining the feature reconstruction and focused pixel recombination, we compare GRFusion with three fusion modes, where mode1 generates the fusion image by $\bm I_{1} \odot \bm M_{1} + \bm I_{2} \odot (\bm 1 - \bm M_{1})$, mode2 merges the source image by $\bm I_{1} \odot (\bm 1 - \bm M_{2}) + \bm I_{2} \odot \bm M_{2}$ and mode 3 produces the fusion result by the full-focus image generation method proposed in this paper. In this process, $\bm M_{i}(i=1,2)$ is the binary map of  focused pixel detection result of source image $\bm I_{i}(i=1,2)$. As boxed regions shown in Fig.\ref{labe14}, the defocused information will be introduced when binary map is estimated incorrectly, and fused results will be not so ideal when only the full-focus image generation method is used. In contrast,  GRFusion can integrate the complementary advantages of mode1, mode2 and mode3. Consequently, the fused results with better performance can be achieved. The above results are also objectively evaluated in Table \ref{table3}.

\subsubsection{Ablation Study on Focus Detection}
\begin{figure}[t!]
	\centering
	\includegraphics[width=3.4in,height=2.2in]{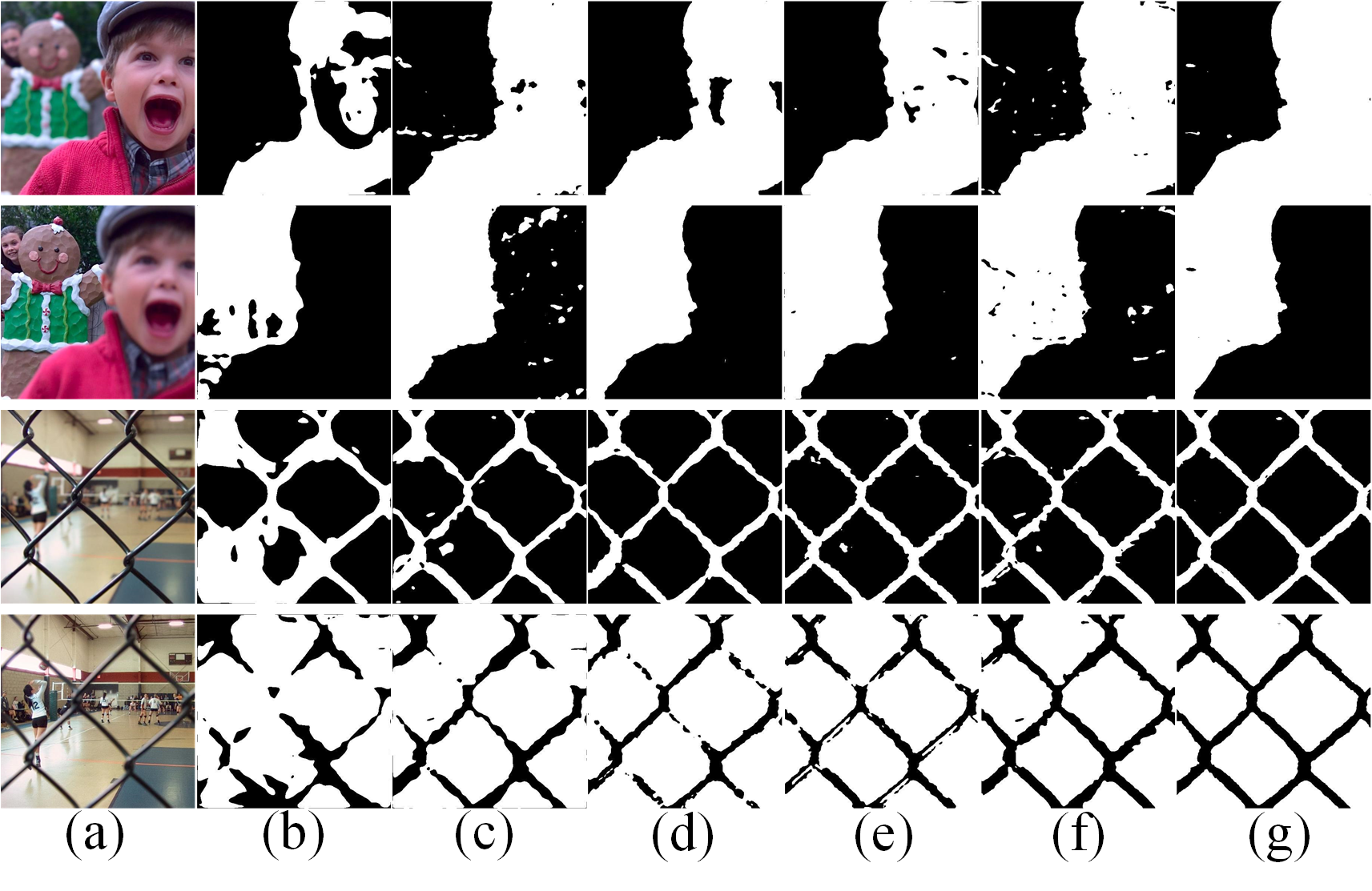}
	\setlength{\abovecaptionskip}{0.cm}
    \caption{Ablation study on focus detection. (a) Source images (b) Baseline (c) Baseline$+$FM$\backslash$MPG (d) Baseline$+$FM$\backslash$Rev (e) Baseline$+$FM (f) Baseline$+$MSFA (g) Baseline$+$FM$+$MSFA.}
    \vspace{-0.4cm}
	\label{labe15}
\end{figure}

\begin{table}
	\caption{Objective evaluation of different methods in FFIG-MFE ablation studies.}
	\label{table4}
	\resizebox{1.0\linewidth}{!}{
		\begin{tabular}{c c c c c|c c c c c}
			\hline
			Baseline&MDEE&WG&EFE&${\bm M_{h}}$&${Q_{MI}}$&${Q_{AB/F}}$&${Q_{CB}}$&${Q_{NCIE}}$&${Q_{SSIM}}$\\
			\hline
			\checkmark & & & & & 0.90407&0.72214&0.70561&0.82991&1.69022\\
			
			\checkmark & & &\checkmark&\checkmark&1.09770&\textcolor{blue}{\bf0.73418}&0.75487&\textcolor{blue}{\bf 0.84118}&1.72432\\
			
			\checkmark&\checkmark& &\checkmark&\checkmark &0.99938&\textcolor{blue}{\bf0.73418}&0.71647&0.83506&1.72021\\
			
			\checkmark&\checkmark&\checkmark& &\checkmark&0.92208&0.72209&0.68758&0.83068&1.69305\\
			
			\checkmark&\checkmark&\checkmark&\checkmark& &\textcolor{blue}{\bf 1.10961}&0.73389&\textcolor{red}{\bf 0.76239}&0.84108&\textcolor{red}{\bf1.72499}\\
			
			\checkmark&\checkmark&\checkmark&\checkmark&\checkmark&\textcolor{red}{\bf 1.11146}& \textcolor{red}{\bf 0.73550}&\textcolor{blue}{\bf 0.75651}&\textcolor{red}{\bf 0.84253}&\textcolor{blue}{\bf1.72491}\\
			\hline
		\end{tabular}
	}
\end{table}

\begin{figure}[t!]
	\centering
	\includegraphics[width=3.3in,height=1.15in]{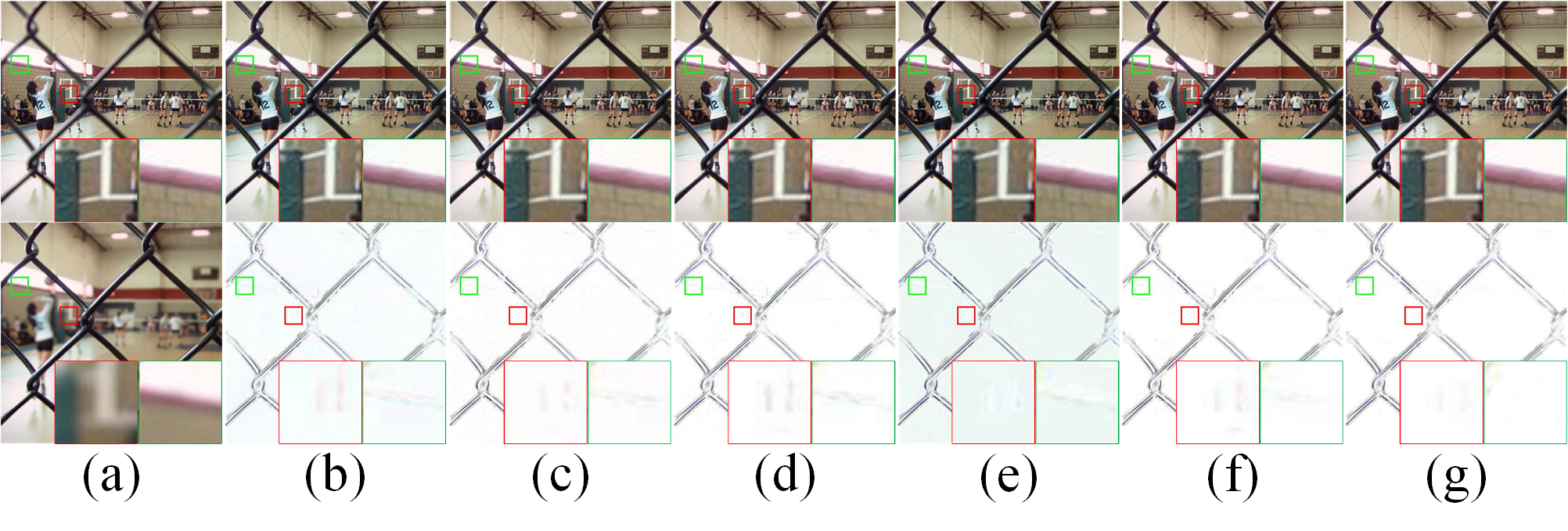}
	\setlength{\abovecaptionskip}{0.cm}
	\caption{Ablation analysis on FFIG-MFE. (a) Source images. (b) Baseline (c) Baseline+EEM+$\bm M_{h}$ (d) Baseline+MDEE+EEM+$\bm M_{h}$ (e) Baseline+MDEE+WG+$\bm M_{h}$ (f) Baseline+MDEE+WG+EEM (g) Baseline+MDEE+WG+EEM+$\bm M_{h}$.}
	\vspace{-0.4cm}
	\label{labe16}
\end{figure}
\begin{figure*}[!t] \centering
	\subfigure[]  {\includegraphics[height=1.0in,width=1.3in,angle=0]{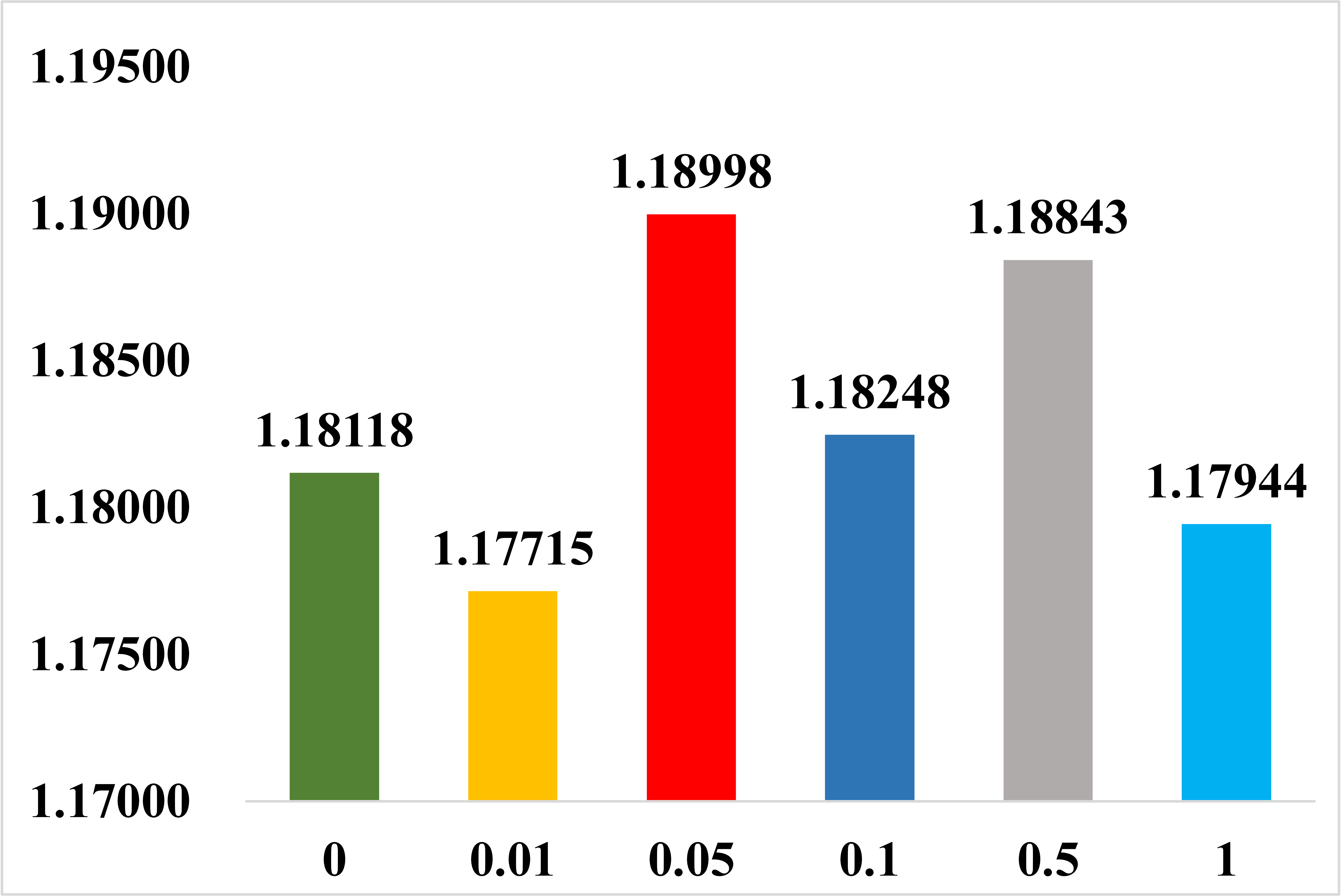}}
	\subfigure[]  {\includegraphics[height=1.0in,width=1.3in,angle=0]{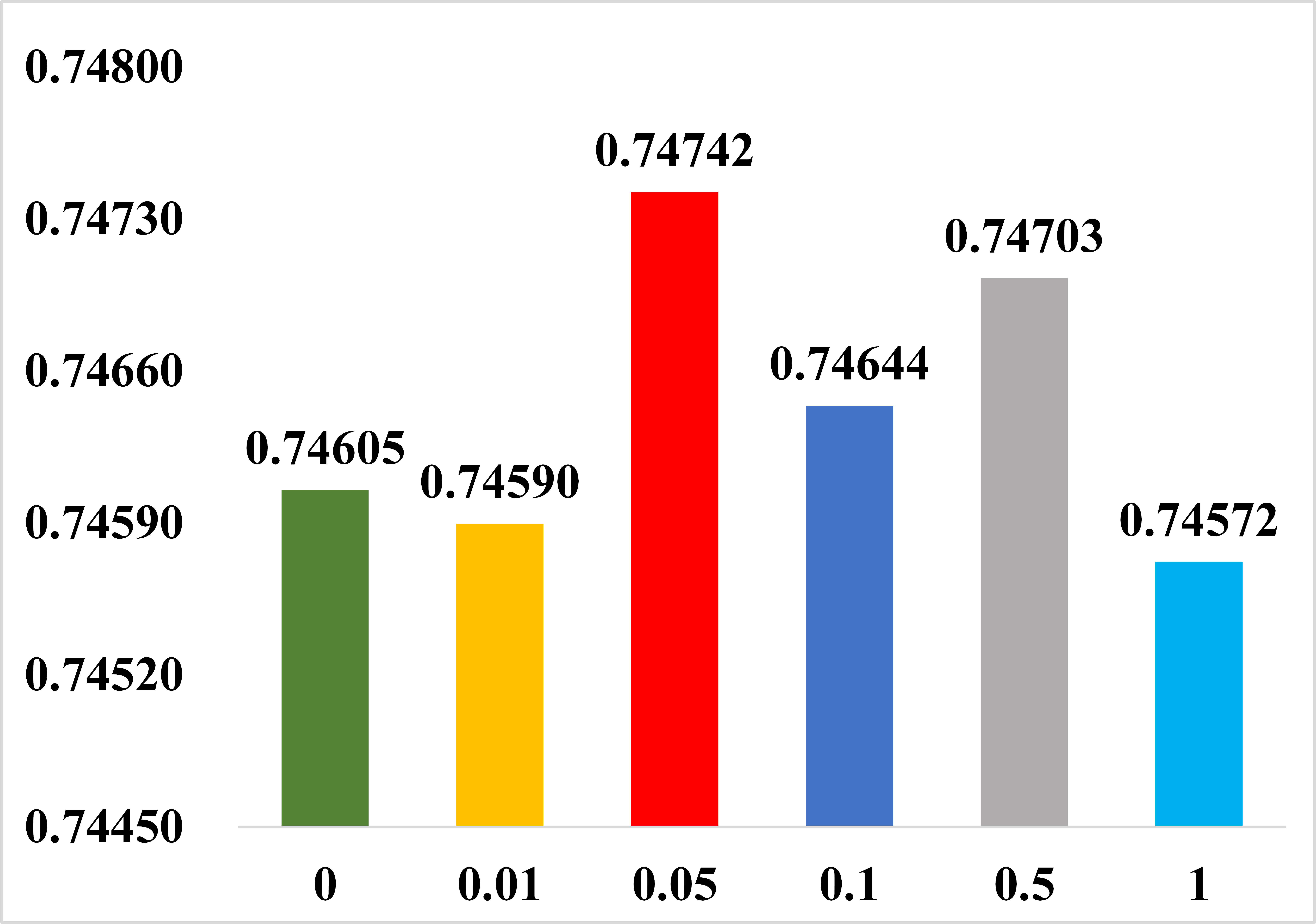}}
	\subfigure[]  {\includegraphics[height=1.0in,width=1.3in,angle=0]{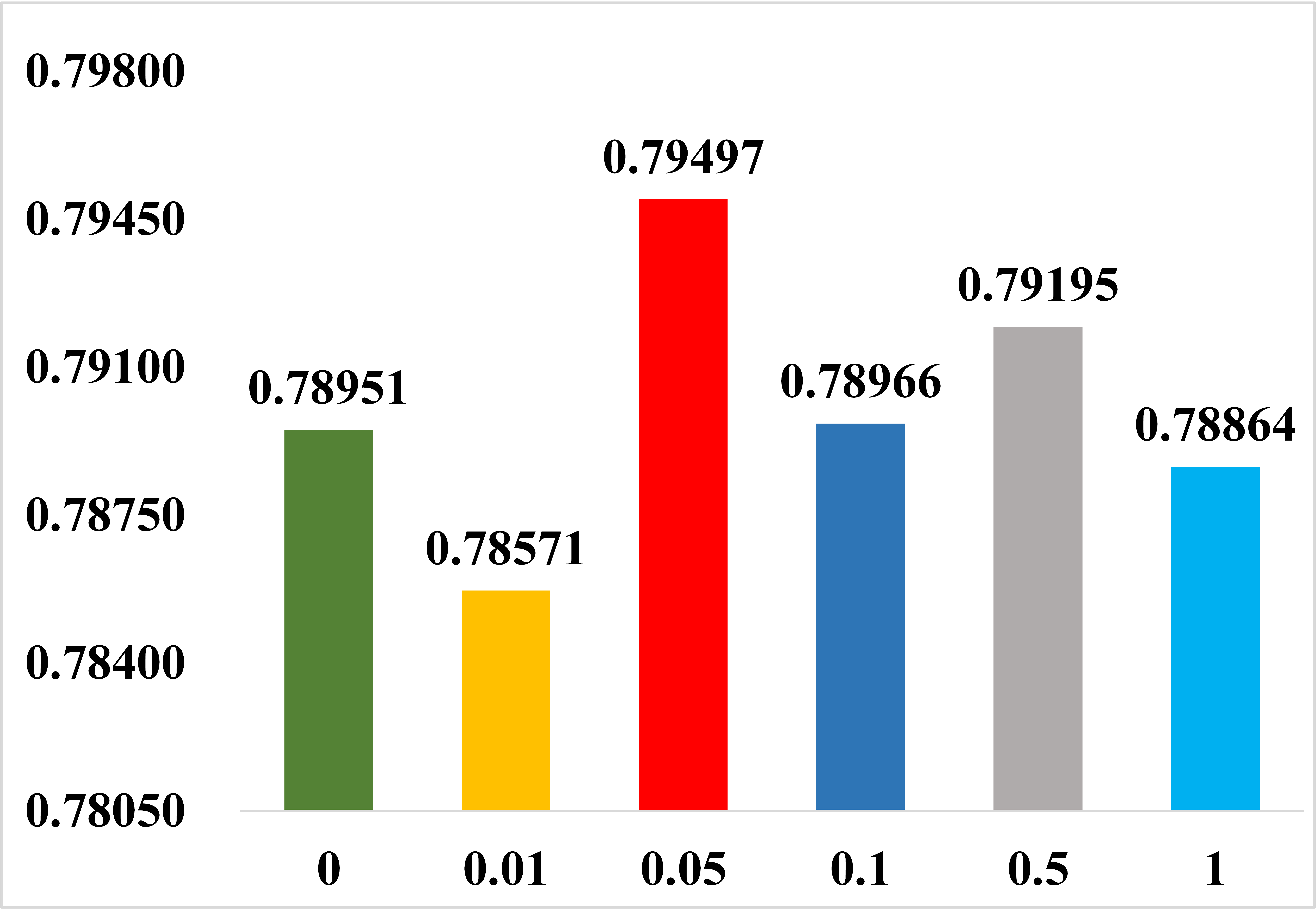}}
	\subfigure[]  {\includegraphics[height=1.0in,width=1.3in,angle=0]{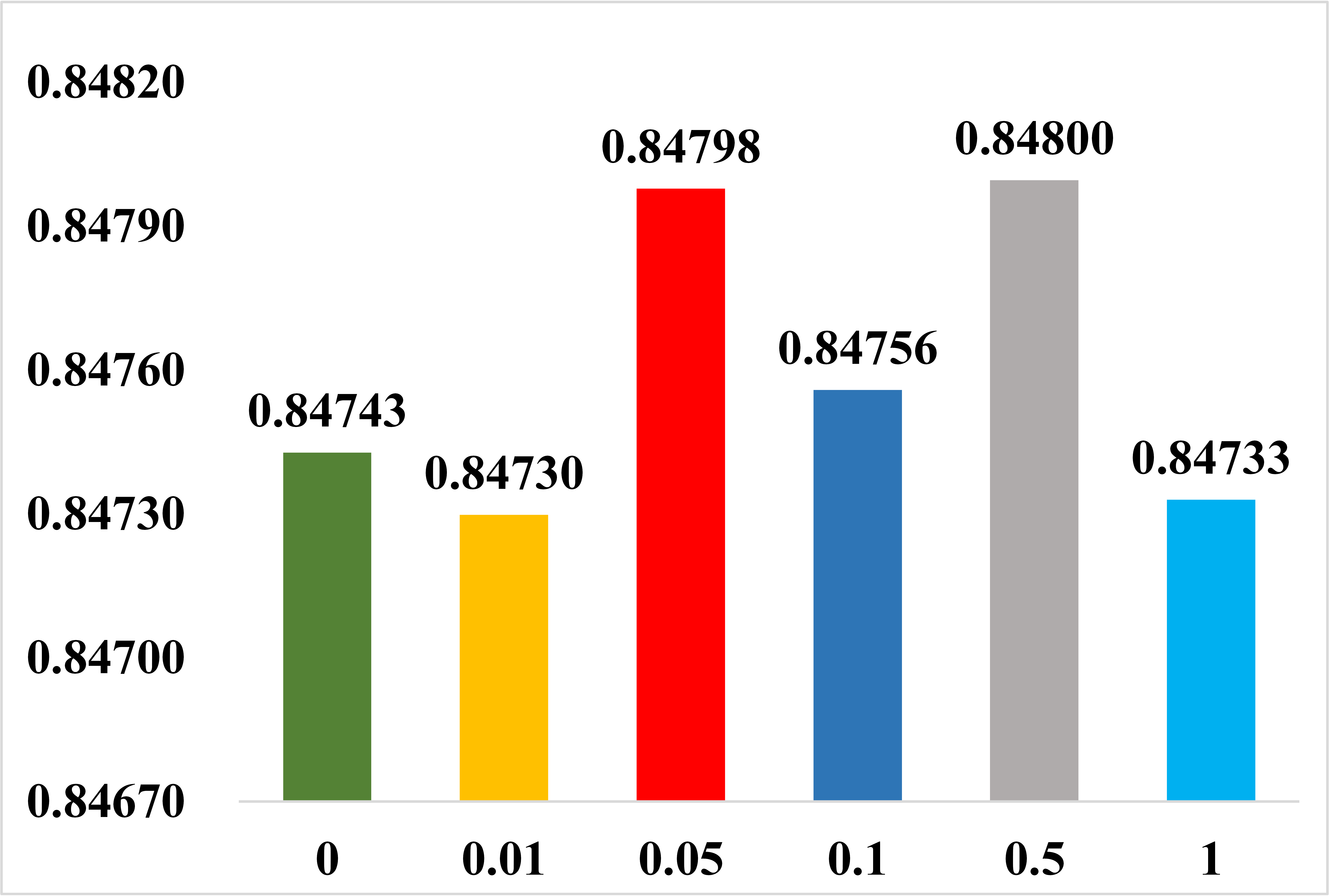}}
	\subfigure[]  {\includegraphics[height=1.0in,width=1.3in,angle=0]{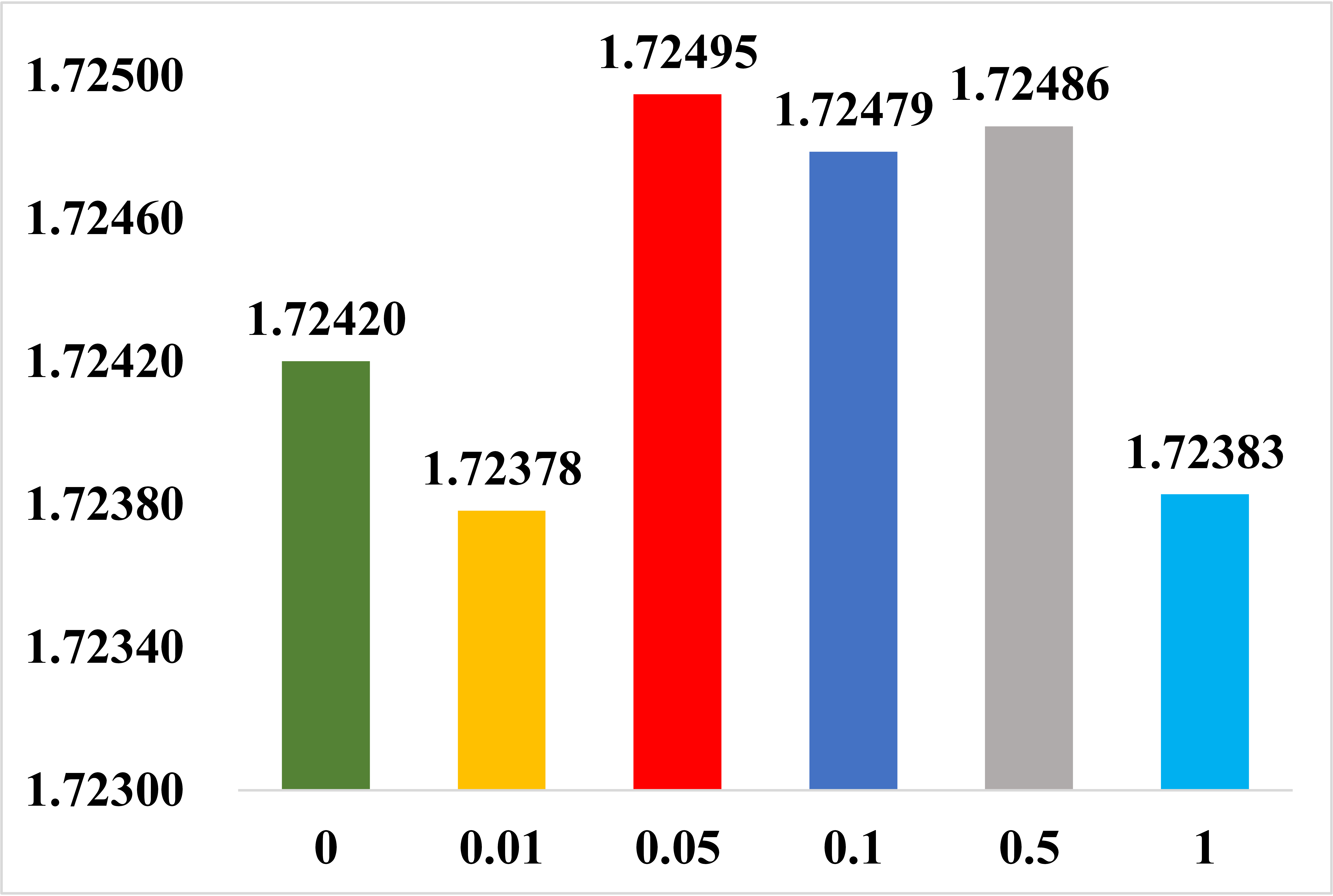}}
	\caption{Objective evaluation when $\lambda$ takes different values. The horizontal axis represents the value of $\lambda$, while the vertical axis represents the objective evaluation result. (a) $Q_{MI}$ (b) $Q_{AB/F}$ (c) $Q_{CB}$ (d) $Q_{NCIE}$ (e) $Q_{SSIM}$.}
	\vspace{-0.4cm}
	\label{labe18}
\end{figure*}
\begin{figure}[!t]
	\centering
	\includegraphics[width=3.4in,height=1.1in]{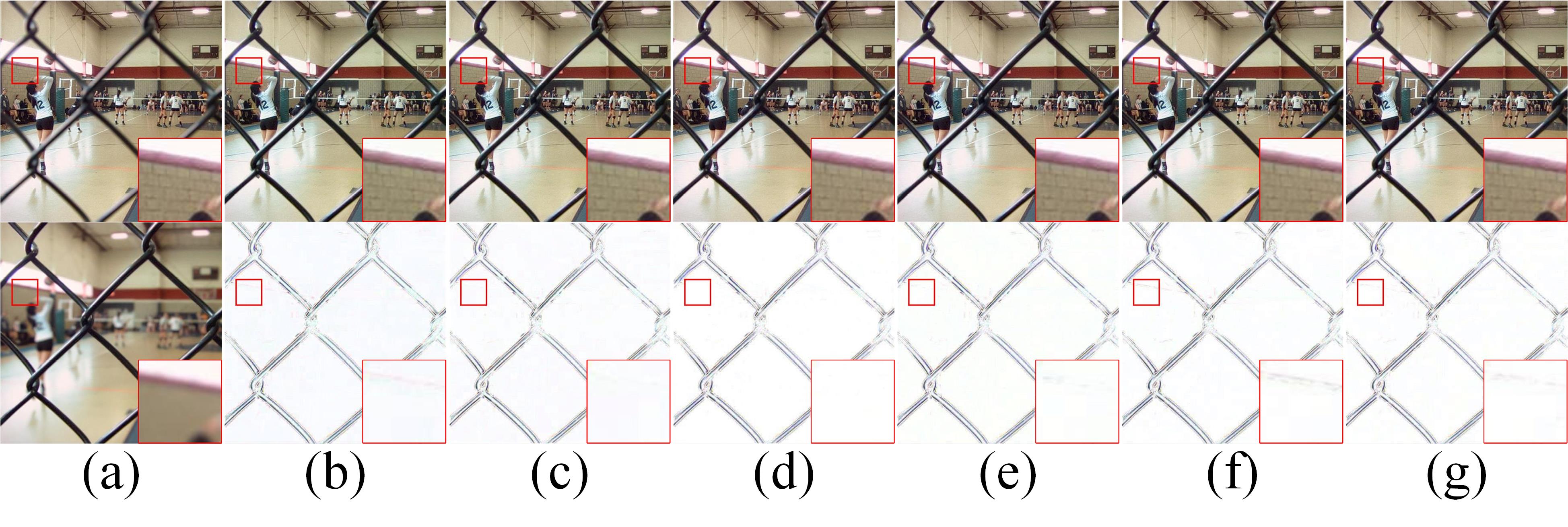}
	\setlength{\abovecaptionskip}{0.cm}
	\caption{The impact of different values of $\lambda$ on fusion performance.(a) Source images (b) $\lambda  = 0$ (c) $\lambda  = 0.01$ (d) $\lambda  = 0.05$ (e) $\lambda  = 0.1$ (f) $\lambda  = 0.5$ (g) $\lambda  = 1$.}
	\vspace{-0.4cm}
	\label{labe17}
\end{figure}
To validate the effectiveness of FM and MSFA in focus property detection, we remove them from HPD respectively. The ablation results are shown in Fig.\ref{labe15}, where ``Baseline'' means that the baseline constructed by removing FM and MSFA from HPD, ``Baseline+FM$\backslash$MPG'' means that the modulation parameter generator is removed from FM, ``Baseline+FM$\backslash$Rev'' means that the reverse operator in Eq.(3) is moved from FM, ``Baseline+FM'' means that FM is added into the Baseline, ``Baseline+MSFA'' means that MSFA is added into the Baseline and ``Baseline+FM+MSFA'' means that both FM and MSFA are added into the Baseline. From these results we can see the effects of different components in HPD, and find that all the components have played a positive role in detecting focused pixels.

\subsubsection{Ablation Analysis on FFIG-MFE}
To verify the effectiveness of different components in FFIG-MFE, ablation experiments are performed for MDEE, EFE, weight generator (WG) and hard pixel information implantation, respectively. In this process, the remaining components of FFIG-MFE after removing MDEE, EFE, WG and hard pixels implantation are used as the baseline. In FFIG-MFE, MDEE is used to extract the edge features from different directions of the source images and then integrate them via WG. In ablation study of MDEE, the Laplace operator is employed to replace MDEE. To investigate the effectiveness of WG and EFE, WG is replaced by addition operation and EFE is replaced with common concatenation operation. In FFIG-MFE, hard pixel information implantation is implemented by Eq.(18). In ablation study of hard pixel information implantation, we directly remove $\bm M_{h}$ from Eq.(18). As seen in Fig.\ref{labe16} and Table \ref{table4}, the performance is optimal when all components are included, and the residual information in the boxed region is also minimal in Fig.\ref{labe16}(g). This demonstrates the effectiveness of each component in FFIG-MFE.
\subsection{Analysis and Selection of $\lambda$}
In our method, there is one critical hyperparameter, \textit{i.e.}, $\lambda$, in loss function.  To search an appropriate value for $\lambda$, we investigate the impact of $\lambda$ when it changes within $[0, 1]$. As shown in Fig.\ref{labe17}, when $\lambda= 0$ and $\lambda  = 0.01$, the fusion result suffers from slight chromatic, and when $\lambda=0.1$, $\lambda=0.5$, $\lambda=1$, there is residual information presenting in the boxed areas. This indicates that not all information of the focused regions has been transferred into the fusion result. When $\lambda=0.05$, the fusion result shows higher quality since there is less residual information remaining in the boxed region. From the visual quality of the fusion results, we can conclude that $\lambda=0.05$ is the best choice. The quantitative evaluation of fusion results when $\lambda$ takes different values is shown in Fig.\ref{labe18}, which further demonstrates superiority  of $\lambda=0.05$ over other settings.
\section{Conclusion}\label{section5}
In this paper, a novel multifocus image fusion method named GRFusion is proposed which combines feature reconstruction with focus pixel recombination. By performing focus property detection of each input source image individually, GRFusion is able to fuse multiple source images simultaneously and avoid information loss caused by alternating fusing strategy effectively.  Meanwhile, the determination of hard pixels is realized based on the inconsistency of all the detection results. To reduce the difficulty of fusion of hard pixels, a full-focus image generation method with multi-directional gradient embedding is proposed. With the generated full-focus image, a hard pixel-guided fusion result construction mechanism is designed, which effectively integrates the respective advantages of the feature reconstruction-based method and the focused pixel recombination-based method. Experiment results and ablation studies demonstrate the effectiveness of the proposed method and each component.
\bibliography{revised-manuscript}
\end{document}